%% file: icml2024.tex
\pdfoutput=1

\documentclass{article}

\usepackage{microtype}
\usepackage{graphicx}
\usepackage{subcaption}
\usepackage{booktabs} 

\usepackage{hyperref}

\usepackage{amsmath}
\usepackage{amssymb}
\usepackage{mathtools}
\usepackage{amsthm}
\input{math_commands.tex}

\usepackage{hyperref}
\usepackage{url}
\usepackage{amssymb}
\usepackage{wrapfig}
\usepackage{graphicx} 
\usepackage{tabularx}
\usepackage{subcaption}
\usepackage{ulem}
\usepackage[ruled,vlined,linesnumbered,noresetcount]{algorithm2e}
\usepackage{bbm}

\usepackage{array}          
\usepackage{booktabs}       
\usepackage{caption}        
\usepackage{multirow}       
\usepackage{adjustbox}
\usepackage[capitalize,noabbrev]{cleveref}

\usepackage[accepted]{icml2024}

\theoremstyle{plain}

\theoremstyle{definition}

\theoremstyle{remark}

\usepackage[textsize=tiny]{todonotes}

\icmltitlerunning{Expert Proximity as Surrogate Rewards for Single Demonstration Imitation Learning}

\begin{document}

\twocolumn[
\icmltitle{Expert Proximity as Surrogate Rewards for\\ Single Demonstration Imitation Learning}



\icmlsetsymbol{equal}{*}

\begin{icmlauthorlist}
\icmlauthor{Chia-Cheng Chiang}{equal,el}
\icmlauthor{Li-Cheng Lan}{equal,ucla}
\icmlauthor{Wei-Fang Sun}{nvaitc}
\icmlauthor{Chien Feng}{el}
\icmlauthor{Cho-Jui Hsieh}{ucla}
\icmlauthor{Chun-Yi Lee}{el}
\end{icmlauthorlist}

\icmlaffiliation{el}{ELSA Lab, Department of Computer Science, National Tsing Hua University, Hsinchu, Taiwan.}
\icmlaffiliation{ucla}{Department of Computer Science, University of California, Los Angeles, CA, USA.}
\icmlaffiliation{nvaitc}{NVIDIA AI Technology Center, NVIDIA Corporation, Santa Clara, CA, USA}

\icmlcorrespondingauthor{Chun-Yi Lee}{cylee@cs.nthu.edu.tw}

\icmlkeywords{Machine Learning, ICML}

\vskip 0.3in
]

\def\K{\mathcal{\tilde{O}}}
\def\O{\mathcal{O}}
\def\Rl2{R_{\text{L2}}}
\def\Rirl{R_{\text{IRL}}}
\def\Rsur{R_{\text{TDIL}}}
\def\Rtot{R_{\text{agg}}}
\def\w{w}
\def\Dphi{D_\phi}
\def\Dphistar{D_{\phi^*}}
\def\sa{s_i}
\def\sb{s_j}
\def\Rweight{\beta}
\def\LD{L_D}
\def\positiveset{B^{+}}
\def\negativeset{B^{-}}
\def\hardnegativeset{B^{-}_{\text{reversed}}}
\def\easynegativeset{B^{-}_{\text{contrastive}}}
\def\defeq{\stackrel{\text{def}}{=}}
\def\1{\mathbbm{1}}
\def\prior{p}



\printAffiliationsAndNotice{\icmlEqualContribution} 

\input{Sections/all.tex}

\bibliography{icml2024}
\bibliographystyle{icml2024}

\newpage
\appendix
\onecolumn
\input{Sections/all_appendix.tex}

\end{document}

%% file: math_commands.tex
\usepackage{amsmath,amsfonts,bm}

\def\eqref#1{equation~\ref{#1}}

\def\1{\bm{1}}

\DeclareMathAlphabet{\mathsfit}{\encodingdefault}{\sfdefault}{m}{sl}
\SetMathAlphabet{\mathsfit}{bold}{\encodingdefault}{\sfdefault}{bx}{n}



%% file: Sections/all.tex
\input{Sections/0_abstract.tex}
\input{Sections/1_introduction.tex}
\input{Sections/2_preliminary}
\input{Sections/3_motivational_example}
\input{Sections/4_methodology.tex}

\input{Sections/5_experiments.tex}

\input{Sections/6_related_works.tex}
\input{Sections/7_conclusion.tex}

\input{Sections/8_acknowledgement.tex}
\input{Sections/9_impact_statement}

%% file: Sections/0_abstract.tex
\begin{abstract}
In this paper, we focus on single-demonstration imitation learning (IL), a practical approach for real-world applications where acquiring multiple expert demonstrations is costly or infeasible and the ground truth reward function is not available. In contrast to typical IL settings with multiple demonstrations, single-demonstration IL involves an agent having access to only one expert trajectory. We highlight the issue of sparse reward signals in this setting and propose to mitigate this issue through our proposed Transition Discriminator-based IL (TDIL) method.
TDIL is an IRL method designed to address reward sparsity by introducing a denser surrogate reward function that considers environmental dynamics. This surrogate reward function encourages the agent to navigate towards states that are proximal to expert states. In practice, TDIL trains a transition discriminator to differentiate between valid and non-valid transitions in a given environment to compute the surrogate rewards.
The experiments demonstrate that TDIL outperforms existing IL approaches and achieves expert-level performance in the single-demonstration IL setting across five widely adopted MuJoCo benchmarks as well as the ``Adroit Door'' robotic environment.
\end{abstract}

%% file: Sections/1_introduction.tex
\section{Introduction}
\label{sec:introduction}

Single-demonstration imitation learning (or simply ``\textit{single-demo IL}'' hereafter) is characterized by an agent having access to only one expert demonstration (i.e., a single expert trajectory). This contrasts with typical IL settings, where multiple demonstrations are available~\citep{ho2016generative, fu2020d4rl}. Both settings allow interactions with the environment during training but lack access to the ground truth reward function, online human feedback, or prior knowledge acquired from analogous tasks. Single-demo IL is a practical paradigm for addressing real-world challenges, as collecting a large number of expert demonstrations is often expensive and sometimes not even feasible, especially in applications such as autonomous robots. 
Consider the training of a surgical robot~\cite{ou2023towards}. In situations where certain surgical procedures are extremely rare, it is possible that only a single expert surgeon's demonstration is available. Similarly, in the context of training an agent for vehicle control~\citep{scheel2022urban}, unique scenarios such as stabilizing the vehicle during a tire blowout, navigating icy roads, or avoiding collisions with objects may have only one or very few demonstrations available. Another example is cooking tutorials on YouTube, where YouTubers typically demonstrate the cooking process only once. As a result, a robotic agent learning from a single-demonstration setting encounters similar challenges in these domains.
However, many IL methods, such as behavior cloning (BC) and most basic inverse reinforcement learning (IRL) methods, face limitations when only a single demonstration is available. BC tends to overfit when few expert demonstrations are provided. For basic IRL methods, the scarcity of expert demonstrations can typically result in a sparse reward situation, which may lead to relatively limited training signals for the agent. This issue of reward sparsity becomes even more pronounced in high-dimensional, continuous environments with randomly initialized positions. In light of these, developing an effective and robust learning mechanism that operates solely with a single demonstration is of considerable importance. Unfortunately, although a few previous methods~\citep{dadashi2021primal,freund2023coupled} exist that can be utilized to address few-demonstration IL, single demo IL remains relatively unexplored and offers opportunities for further advancing contemporary IL approaches.

To confront the single-demo IL paradigm, this study proposes an IRL method with a denser reward function, termed Transition Discriminator-based IL (TDIL). TDIL increases the density of obtainable reward signals in the IRL setting while accounting for environmental dynamics to ensure reasonable agent behavior. A motivational example illustrating this concept is provided in Fig.~\ref{fig:motivation}~(a).
If an agent finds itself in a cell that allows for a direct transition to an expert state (e.g., the green arrows in Fig.~\ref{fig:motivation}~(a)), the most reasonable action is to facilitate this transition. Based on this concept, TDIL derives a surrogate reward function that rewards the agent for moving toward states close to expert states. Although there exist other dense reward IL methods (e.g., PWIL~\citep{dadashi2021primal} and FISH~\citep{haldar2023teach}) designed to guide the agent to states that are close to expert states, the distance metrics they employed, such as the Euclidean (L2) or the cosine distance metrics, are not theoretically sound. For example, in Fig.~\ref{fig:motivation}~(a), if the L2 distance is adopted, two adjacent grid cells would be considered close to each other even if a barrier exists between them. This could potentially lead the agent to cells that are either infeasible or unable to reach expert states in an efficient manner. In contrast, TDIL takes environmental dynamics into consideration by leveraging a well-trained \textit{transition discriminator}, which adopts a training objective aimed at distinguishing between valid and non-valid transitions regarding two states' reachability in a given environment. As a result, TDIL is able to construct a more reasonable and denser reward function (e.g., Fig.~\ref{fig:motivation}~(d)) for guiding the agent back to expert states in the single-demo IL setting.

To validate the efficacy of TDIL, we perform comprehensive experiments on five widely adopted MuJoCo benchmarks~\citep{todorov2012mujoco}, aligning with most prior IL research, as well as the ``Adroit Door'' environment~\citep{rajeswaran2017learning} in the Gymnasium-Robotics collection~\citep{gymnasium_robotics2023github}. The experimental evidence reveals that TDIL delivers exceptional performance, matches expert-level results on these benchmarks, and outperforms existing IL approaches. Moreover, another key insight from our experiments is the significant correlation between the derived reward signals and the inaccessible ground truth reward signals. This correlation offers a practical solution for blind model selection by selecting a checkpoint without the help of the ground truth reward function. This differentiates TDIL from prior work that relied on environment rewards at test time for early termination or optimal model selection, which are assumptions that are impractical in the general IL context. The main contributions 
are summarized as follows:
\vspace{-1.5em}
\begin{enumerate}
    \item We highlight the limitations of previous IL methods under the challenging single-demo IL setting. These methods may produce sparse reward signals or sometimes even overlook the dynamics of the environment.
    \item We introduce a novel TDIL algorithm, which utilizes a dense and dynamics-aware surrogate reward function.
    \item We validate that our surrogate reward function is effective for blind model selection scenarios without requiring access to the ground truth reward function.
\end{enumerate}

%% file: Sections/2_preliminary.tex
\section{Preliminary}
\label{sec:background}

\textbf{Reinforcement learning (RL).}~An MDP is typically formalized as a tuple $\langle\mathcal{S}, \mathcal{A}, P, R, p_{0}\rangle$, where $\mathcal{S}$ represents the state space, $\mathcal{A}$ the action space, $P: \mathcal{S} \times \mathcal{A} \times \mathcal{S} \to \mathbb{R}$ the transition function, $R(s, a): \mathcal{S} \times \mathcal{A} \to \mathbb{R}$ the reward function, and $p_{0}(s_0)$ the distribution of the initial state $s_0$. The transition function $P(s_{t+1} | s_{t}, a_{t})$ specifies the probability of transitioning to state $s_{t+1}$ upon taking action $a_t$ in state $s_t$. Within this MDP, a trajectory $\tau$ is defined as a sequence of states and actions $[s_0, a_0, s_1, a_1, \ldots, s_{T}, a_T]$, where $s_0$ is sampled from the distribution $p_0$, and $s_{t+1}$ is the resulting state after taking action $a_t$ in state $s_t$. The objective of an RL policy $\pi(a|s,\theta)$ is to learn a set of parameters $\theta$ that maximizes the expected total return $\mathbb{E}_{\tau \sim p(\tau |\theta)}\sum_{t=0}^{T} R(s_t, a_t)$. 
\textbf{Single-demo IL.}~During the training of single-demo IL settings, the agent can interact with the environment. However, it does not have access to the reward function $R$. Instead, the agent is given a trajectory $\tau_e = [s^e_0, a^e_0, s^e_1, a^e_1, \dots, s^e_T, a^e_T]$ generated by an expert policy $\pi_e$ in the same environment as a hint of the reward $R$. As a result, the goal of single-demo IL is to train a policy that can converge to the expert demonstration even when initiated from a different initial state $s_0$, and faithfully follow the expert actions when within the support of the demonstration. After training, the performance of it is evaluated by the ground truth reward function $R$.

\textbf{Inverse reinforcement learning (IRL).}~IRL methods constitute a type of IL that aims to learn or infer the reward function based on provided demonstrations. \cite{levine2018reinforcement} demonstrated that the objective of IRL is to learn a Conditional Probability Distribution (CPD) denoted as $p(\O_t=1|s_t, a_t)$. In this expression, the optimal indicator $\O_t$ serves as a binary random variable that indicates whether the time step $t$ is optimal. Specifically, in the context of IRL, $\O_t=1$ if the $(s_t, a_t)$ pair is present in an expert trajectory. Furthermore, the CPD $p(\O_t=1|s_t, a_t)$ can be marginalized to form $p(\O_t=1|s_t) = \int_{\mathcal{A}} \prior(a_t|s_t)p(\O_t=1|s_t, a_t)da_t$. By assuming the action prior $\prior(a|s)$ produces the expert actions in the expert states, 
$p(\O_t=1|s_t)=1$ if and only if $s_t$ is an expert state. The assumption can be ensured through BC or GAIL.
In the following sections, we slightly abuse notation (i.e., dropping $=1$) as in~\citet{levine2018reinforcement} for the sake of conciseness by expressing $p(\O=1)$ as $p(\O)$.

%% file: Sections/3_motivational_example.tex
\section{Analysis on the Sparse Reward Issue}
\label{sec:motivation}

\input{Figures/motivation}
\input{Figures/converge}

To examine the sparse reward issue in single-demo IL scenarios, we design a 2D grid-world environment and compare the reward functions, learned optimal policies, and training curves across different IL methods.
Fig.~\ref{fig:motivation}~(a) illustrates the grid-world environment, where the circled triangle symbol denotes the initial state of the expert, while the red lines represent barriers that obstruct certain paths. The blue arrows trace the path of the expert's demonstration as it progresses toward the goal state, which is marked by a yellow flag.

We investigate three distinct approaches for defining the reward functions: (1) the basic IRL reward (i.e.,~\citet{ho2016generative}), (2) the L2 distance reward, and (3) the TDIL reward. The basic IRL reward only provides rewards to the agent when it performs expert demonstrations (i.e., the blue arrows).
The L2 distance reward provides rewards according to the L2 distance between the current state-action of the agent and that of the expert. This reward represents methods that use geometric distance to measure the spatial disparity between the two
(e.g, PWIL~\citep{dadashi2021primal}, FISH~\citep{haldar2023teach}, and ROT~\citep{haldar2023watch}).
Finally, our TDIL reward considers the transitions to expert state reasonable (i.e., the green arrows) and provides rewards to them. The corresponding reward functions of these three cases are visualized in Figs.~\ref{fig:motivation}~(b)-(d).
Subsequently, we train Soft Actor-Critic (SAC)~\citep{haarnoja2018soft} agents under a uniform initial state distribution $p_0(s_0)$, using each of these reward functions.
The critic in SAC facilitates the propagation of reward signals to cells that do not provide any rewards.
In Figs.~\ref{fig:motivation}~(f)-(h), the agents all start at the bottom-left corner, with the learned policy at each state indicated by an arrow and the trajectories of the agents highlighted with a grey background. The policy learned through BC is included in Fig.~\ref{fig:motivation}~(e) for comparison. The training curves 
are presented in Fig.~\ref{fig:converge}, indicating the steps required to reach the goal from the bottom-left corner.

According to the above setup, it can be observed that in this single-demo IL scenario, the basic IRL method results in a sparse reward function (i.e., Fig.~\ref{fig:motivation}~(b)), and necessitates more training steps to converge, as illustrated by the orange curve in Fig.~\ref{fig:converge}. This issue becomes more pronounced in high-dimensional environments with a continuous state space, where the expert trajectory may represent a low-dimensional manifold with measure zero. 
In such scenarios, basic IRL methods that aimed to minimize the f-divergences 
between the agent and expert state distributions
often encounter convergence challenges. These challenges stem from the difficulties of matching two manifolds with significantly 
different dimensions~\citep{arjovsky2017towards}.

For the L2 distance reward, although it leads to a more densely defined reward function as illustrated in Fig.~\ref{fig:motivation}~(c), the learned policy tends to become trapped in the states above the goal state, which are obstructed by red barriers, as depicted in Fig.~\ref{fig:motivation}~(g). This can be attributed to the reliance on distance measures that do not adequately reflect the state transition dynamics inherent in the underlying MDP. Despite the proximity to the expert support based on the L2 distance, these states are actually far separated when the influence of state transition dynamics is taken into account. 
Although this issue can be alleviated by a better state representation, tailoring such representations for different environments can be a challenging task.
This process may require domain-specific knowledge and may not be generalized well to unseen environments.
Moreover, even with a perfect representation, geometric distances such as L2 and cosine similarity may not adequately capture the dynamics of certain environments. Consider the example of driving a car on a highway. Geometric distances treat moving forward and backward as equally valid options, which is reflected in the symmetry property ($L2(f(s_i), f(s_j)) = L2(f(s_j), f(s_i))$, where $f$ is the representation function). However, in reality, car movement on a highway is asymmetric. While it is possible to move forward freely, attempting to move backward at high speeds is dangerous or even impossible.
As a result, using geometric distance to define a dense reward function may not suit environments with complex transition functions.

Finally, the proposed TDIL method results in a denser surrogate reward function, while considering the state transition dynamics of the MDP, as shown in Fig.~\ref{fig:motivation}~(d).
TDIL achieves this by providing rewards to the state-action pairs that return to expert proximity states (including states highlighted by yellow in Fig.~\ref{fig:expertprox}).
While these state-action pairs may not necessarily correspond to the shortest path to the goal state, we posit that returning to the expert support first is the most conservative decision for out-of-distribution states (i.e., Fig.~\ref{fig:motivation}~(h)), given the absence of the ground truth reward function.
Moreover, introducing incentives for the agent to navigate to the expert support can lead to an accelerated convergence speed, as indicated by the green curve in Fig.~\ref{fig:converge}.

By employing TDIL to define a denser surrogate reward function, the sparse reward issue is mitigated when compared to using the basic IRL methods. In high-dimensional environments, even when provided with an expert trajectory within a low-dimensional manifold with measure zero,
TDIL can offer the potential to define a manifold with a higher dimension and a non-zero measure, thereby improving the stability of IRL methods in matching the agent and expert state distributions.
On the other hand, in comparison to the methods that use geometry distance, TDIL exhibits the ability to assign more reasonable rewards due to its awareness of state transition dynamics, as evidenced by the states above the goal state in Figs.~\ref{fig:motivation}~(c)-(d). 
This feature prevents the agent from being trapped in states with high L2 distance rewards that are distant from the expert support. As a result, TDIL holds the potential to enhance training efficiency by adopting a dense and dynamics-aware surrogate reward function, which enables the agent to propagate reward signals back across the states in the environment.

%% file: Figures/motivation.tex
\begin{figure*}[t]
\includegraphics[width=\linewidth]{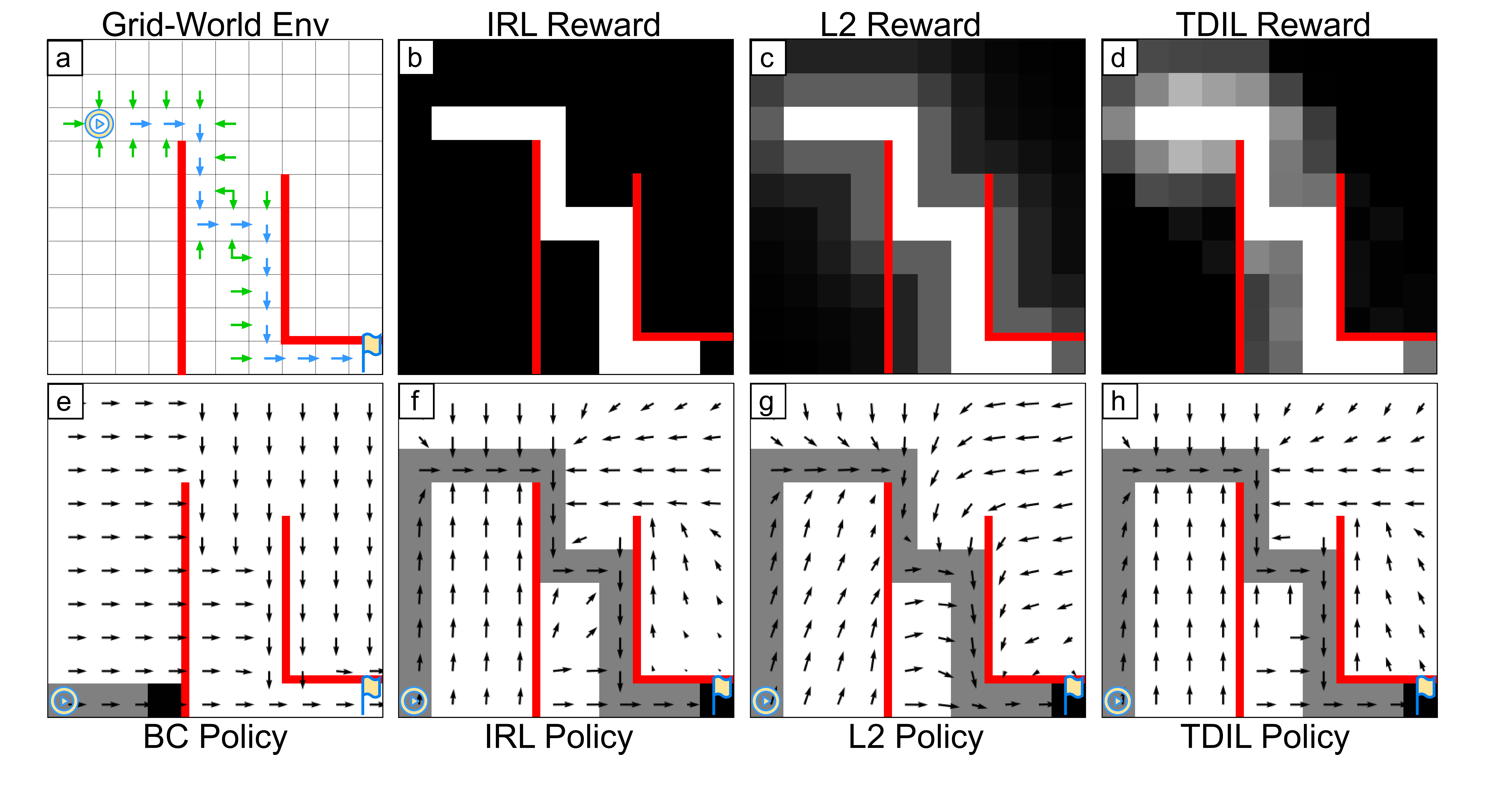}
\centering
\vspace{-4em}
\caption{
A motivational grid-world example for comparing different IL methods trained with the single-demo IL setting. (a) depicts the expert's demonstration, denoted by blue arrows, while red lines represent impassable barriers, reflecting environmental dynamics. The green arrows symbolize the state-action pairs that are one step directed toward the expert states. Subfigures (b)-(d) present reward signals calculated through various methods: (b) using the basic IRL method (i.e., GAIL~\citep{ho2016generative}), (c) based on the L2 distance between the agent's and the expert's state-action pairs, and (d) through our proposed TDIL. Finally, subfigures (e)-(h) illustrate the actions calculated by averaging the directions represented by the logits for the discrete actions from the learned policy at distinct grid locations.
}
\vspace{-1.5em}
\label{fig:motivation}
\end{figure*}

%% file: Figures/converge.tex
\begin{figure}[t]
\includegraphics[width=1\linewidth]{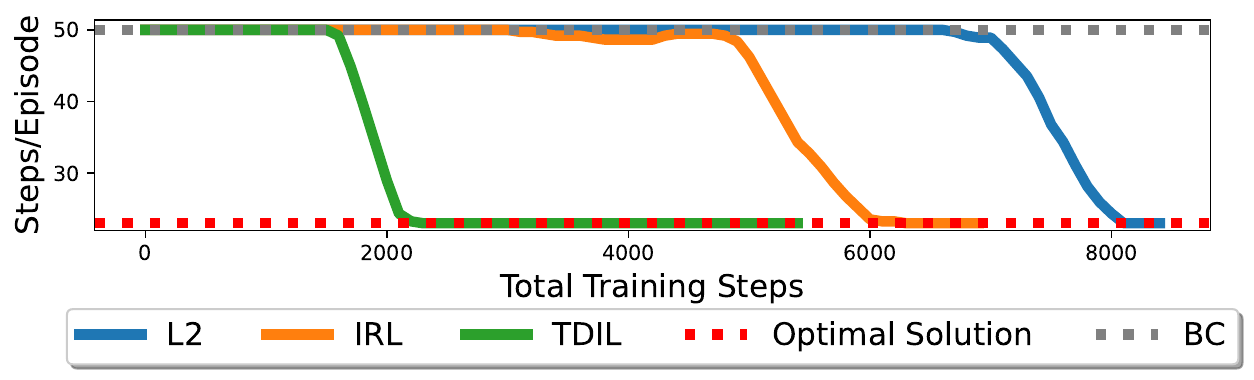}
\centering
\caption{The total steps per episode required for each agent to reach the goal in the grid-world, with a fixed limit of 50 steps. Lower values for ``Steps per episode'' indicate better efficiency.
}
\vspace{-1em}
\label{fig:converge}
\end{figure}

%% file: Sections/4_methodology.tex
\section{Methodology}
\label{sec:methodology}
Section~\ref{subsec::expert_proximity_and_surrogate_rewards} introduces the expert reachability indicator to establish the concept of expert proximity. Subsequently, a denser, dynamics-aware surrogate reward function is formally defined.
However, the computation of surrogate rewards is intractable.
To address this challenge, Section~\ref{subsec::approximating_the_surrogate_reward} presents an approximation for the surrogate reward function.
Section~\ref{subsec::training_the_transition_discriminator} describes the steps to realize this approximation through the use of a transition discriminator.
Building upon these concepts, Section~\ref{subsec::tdil}, introduces a practical algorithm named TDIL, which is designed to facilitate concurrent training of the transition discriminator and the agent. Moreover, Section~\ref{subsec::TD_selection} discusses the advantage of using state based transition discriminator over state-action based one. Finally, Section~\ref{subsec::relative_rewards_for_blind_model_selection} explores the capability of using TDIL for blind model selection, which enables the selection of a proper
checkpoint without relying on ground truth rewards.

\input{Sections/4_1_expert_proximity}
\input{Sections/4_2_approximate_surrogate_reward}
\input{Sections/4_3_trainingTD}
\input{Sections/4_4_TDIL_algo}
\input{Sections/4_5_TD_selection}
\input{Sections/4_6_relative_reward}

%% file: Sections/4_1_expert_proximity.tex
\subsection{Expert Proximity and Surrogate Rewards}
\label{subsec::expert_proximity_and_surrogate_rewards}

Based on the optimal indicator $\O_t$  described in Section~\ref{sec:background}, we define the expert reachability indicator $\K_t$, which identifies the state-action pairs capable of returning to an expert state (i.e., green arrows in Fig.~\ref{fig:motivation} (a)). 
For a given state-action pair $(s_t, a_t)$, we define $p(\K_t|s_t, a_t)$ based on $\O_t$, with $p(\K_t|s_t, a_t)$ indicating the probability of reaching an expert state by selecting action $a_t$ in state $s_t$.
Taking Fig.~\ref{fig:motivation} (a) as an example, if $(s_t, a_t)$ is one of the green or blue arrows, $p(\K_t|s_t, a_t) = 1$ as the agent can reach an expert state from state $s_t$. Formally, we define $p(\K_t|s_t, a_t)$ as follows:
\input{Equations/definition_of_o_tile}%
where $P$ is the state transition function of the MDP and $p(\O_{t+1}|s_{t+1})$ is the probability of $s_{t+1}$ being an expert state. 
Given the action prior $\prior(a|s)$ described in Section~\ref{sec:background}, $p(\K_t|s_t)$ can be derived by marginalizing $p(\K_t|s_t, a_t)$ as:
\input{Equations/def_of_P_O_S}Based on $p(\K_t|s_t)$, we define \textit{expert proximity} as the set of states capable of transitioning to expert states within a single action. In other words, a state $s_t$ is in expert proximity if and only if $\K_t=1$.
Nevertheless, calculating $p(\O_{t+1}|s_{t+1})$ requires access to the ground truth expert support, which is unavailable in general. Fortunately, in the context of single-demonstration IL settings, it is possible to derive the probability $\hat{p}(\K_t|s_t)$ of reaching expert states as follows:
\input{Equations/expert_prox_to_sample}%
where $s^e_i$ denotes the $i$-th state in the expert demonstration, and $N$ represents the total number of expert states.
Finally, we define our surrogate reward function $\Rsur(s_t,a_t)$ as:
\begin{equation}
\begin{aligned}
\Rsur(s_t,a_t) &\defeq \mathbb{E}_{s_{t+1}\sim P(s_{t+1}|s_t,a_t)}\Big[\hat{p}(\K_{t+1}|s_{t+1}) \Big],
\label{eq:surrogate_reward}
\end{aligned}
\end{equation}
which assigns positive surrogate rewards when transitioning to states in expert proximity. This denser reward function also facilitates the propagation of rewards to earlier states in the agent's trajectory and, therefore, can potentially improve its training speed and efficiency as discussed in Section~\ref{sec:motivation}.

%% file: Equations/definition_of_o_tile.tex
\begin{equation}
\begin{aligned}
p(\K_t|s_t,a_t) &\defeq \int_{\mathcal{S}} P(s_{t+1}|s_t,a_t) p(\O_{t+1}|s_{t+1}) ds_{t+1}, \\
\end{aligned}
\label{eq:proximity_indicator_st_at}
\end{equation}

%% file: Equations/def_of_P_O_S.tex
\begin{equation}
\begin{aligned}
&p(\K_t|s_t) = \int_{\mathcal{A}} \prior(a_t|s_t) p(\K_t|s_t,a_t) da_t \\
&= \int_{\mathcal{A}} \prior(a_t|s_t) \int_{\mathcal{S}} P(s_{t+1}|s_t,a_t) p(\O_{t+1}|s_{t+1}) ds_{t+1} da_t \\
&= \int_{\mathcal{S}} p(\O_{t+1}|s_{t+1}) \int_{\mathcal{A}} \prior(a_t|s_t) P(s_{t+1}|s_t,a_t) da_t ds_{t+1} .\\
\end{aligned}
\label{eq:proximity_indicator_st}
\end{equation}

%% file: Equations/expert_prox_to_sample.tex
\begin{equation}
\begin{aligned}
\hat{p}(\K_t| s_t) &= \sum_{i=0}^{N} p(\O_{t+1}|s^e_i) \int_{\mathcal{A}} \prior(a_t|s_t) P(s^e_i|s_t,a_t) da_t \\
&= \sum_{i=0}^{N} \int_{\mathcal{A}} \prior(a_t|s_t) P(s^e_i|s_t,a_t)  da_t,
\end{aligned}
\label{eq:expert_prox_to_sample}
\end{equation}

%% file: Sections/4_2_approximate_surrogate_reward.tex
\subsection{Approximating the Surrogate Reward}
\label{subsec::approximating_the_surrogate_reward}

The computation of $\Rsur$ involves an intractable integration term $\int_{\mathcal{A}} \prior(a_{t}|s_{t}) P(s^e_i|s_{t},a_{t}) da_{t}$ as specified in Eq.~(\ref{eq:expert_prox_to_sample}). To circumvent the complexity introduced by this intractable term, we assume that the action prior $\prior(a_{t}|s_{t})$ is optimal and deterministic in the states that are in expert proximity. This enables us to reformulate the intractable term as follows:
\begin{equation}
\begin{aligned}
\int_{\mathcal{A}} \prior(a_{t}|s_{t}) P(s^e_i|s_{t},a_{t}) da_{t} = \max_{a} P(s^e_i|s_{t},a).
\end{aligned}
\label{eq:convert}
\end{equation}
Eq.~(\ref{eq:convert}) determines an agent's capability to transition from $s_t$ to $s^e_i$, which cannot be computed directly due to the inaccessibility of state transition dynamics. As a result, we train a transition discriminator $\Dphi(s_i, s_j)$ to approximate the state transition dynamics, which determines whether a given state $s_i$ can reach another state $s_j$ within a single timestep.
For example, for any tuple $(s_t, a_t, s_{t+1})$ in the replay buffer, $\Dphi(s_t, s_{t+1})$ should return $1$ since the tuple evidences the reachability. The optimal transition discriminator $\Dphistar(s_i, s_j)$ can be formally defined as follows:
\begin{equation}
\begin{aligned}
\Dphistar(\sa,\sb) &\defeq \max_{a_i}\1[P(\sb|\sa,a_i) > 0].
\end{aligned}
\label{eq:dphi}
\end{equation}
The surrogate rewards $\Rsur$ can then be approximated as:
\begin{equation}
\begin{aligned}
\Rsur(\sa,\sb) &\approx \mathbb{E}_{s_{t+1}\sim p(s_{t+1}|s_t,a_t)}\Big[ \sum_{i=0}^{N} \Dphi(s_{t+1},s^e_i) \Big],
\end{aligned}
\label{eq:approx_r_tdil}
\end{equation}
where the workflow of approximating $\Rsur$ through the use of a given transition discriminator $\Dphi$ is depicted in Fig.~\ref{fig:transition_discriminator}.

%% file: Sections/4_3_trainingTD.tex
\subsection{Training the Transition Discriminator}
\label{subsec::training_the_transition_discriminator}

To train $\Dphi$, we optimize it using maximum likelihood training, with the binary cross-entropy loss $\LD$ defined as:
\input{Equations/binary_cross_entropy}%
where $\alpha \in (0,1)$ is a balancing coefficient, $\positiveset$ is the set of positive samples, and $\negativeset$ is the set of negative samples. In practice, we choose the set of positive samples $\positiveset$ as:
\input{Equations/positive_sample}%
where $B$ is the replay buffer. 
For negative samples $\negativeset$, we choose the union of the set of contrastive samples (i.e., easy negative samples) $\easynegativeset$ and the set of reversed transition samples (i.e., hard negative samples) $\hardnegativeset$ as:
\input{Equations/negative_sample}The positive samples are taken from valid transitions collected by the agent. For the negative samples, we assume that two randomly sampled states seldom represent a valid transition and that the majority of reversed transitions are likely to be invalid. Based on this assumption, the transition discriminator is trained with millions of positive and negative state transitions gathered through the agent's interaction with the environment during training. This method mitigates the likelihood of overfitting compared to the previous work~\cite{ho2016generative}, which uses only expert demonstrations as positive examples for training the discriminator.

%% file: Equations/binary_cross_entropy.tex
\begin{equation}
\begin{aligned}
\LD=&-\Big( \alpha \mathbb{E}_{(\sa,\sb)\sim \positiveset}\big[\log{(\Dphi(\sa,\sb))}\big]+\\
&(1-\alpha)\mathbb{E}_{(\sa,\sb)\sim \negativeset}\big[\log{(1-\Dphi(\sa,\sb))}\big] \Big),
\label{eq:bce}
\end{aligned}
\end{equation}

%% file: Equations/positive_sample.tex
\begin{equation}
    \positiveset = \{ (s,s') \ |\ (s, a, s') \in B \},
    \label{eq:ps}
\end{equation}

%% file: Equations/negative_sample.tex
\begin{equation}
\begin{aligned}
    &\negativeset = \easynegativeset \cup \hardnegativeset, \text{where}\\
    &\easynegativeset = \{ (\sa, \sb) \ |\ (\sa, a_i, s_{i+1}), (\sb, a_j, s_{j+1}) \in B \},\\
    &\hardnegativeset = \{ (s',s) \ |\ (s, a, s') \in B \}.
    \label{eq:ns}
\end{aligned}
\end{equation}

%% file: Sections/4_4_TDIL_algo.tex
\subsection{The TDIL Algorithm}
\label{subsec::tdil}

\input{Figures/framework}

Fig.~\ref{fig:overview} presents an overview of the proposed TDIL
algorithm,
which involves repeating the following four steps.
First, the agent interacts with the environment and stores the collected transitions in the replay buffer. These transitions are then utilized to update the transition discriminator according to Eq.~(\ref{eq:bce}) in the second step. In the third step, a batch of transitions is sampled
from the replay buffer to calculate the aggregated reward $\Rtot$, which is defined as the following:
\begin{equation}
\begin{aligned}
\Rtot(s_t,a_t) &\defeq \Rweight \Rirl(s_t,a_t) + (1-\Rweight)\Rsur(s_t,a_t),
\label{eq:total_reward}
\end{aligned}
\end{equation}
where $\Rweight$ is a hyperparameter for balancing between the two rewards. The aggregated rewards $\Rtot$ combine the basic IRL rewards $\Rirl$, which ensures optimality on the expert and expert proximity states, and the proposed 
$\Rsur$, which incentivizes the agent to navigate towards states that are in expert proximity. In the fourth step, the sampled transitions and the aggregated rewards are utilized to train a SAC agent. This iterative process facilitates the concurrent training of the transition discriminator and the agent.
For the pseudocode and additional details of the proposed TDIL algorithm, please refer to Section~\ref{apx:algo_detail}.
In practice, the basic IRL rewards $\Rirl$ are calculated using GAIL. In addition, an ablation study that explores different choices of $\beta$ is presented in Section~\ref{apx:experiments:beta}. Furthermore, we explore another variant where $\Rirl$ (i.e., $\beta=0$) is removed, and an additional BC loss is employed to train the policy to ensure the optimality on expert states, as described in Section~\ref{sec:background}. 

%% file: Figures/framework.tex
\begin{figure*}
	\begin{minipage}{0.36\linewidth}
	\centering
    \vspace{1.5em}
    \footnotesize
    \includegraphics[width=\linewidth]{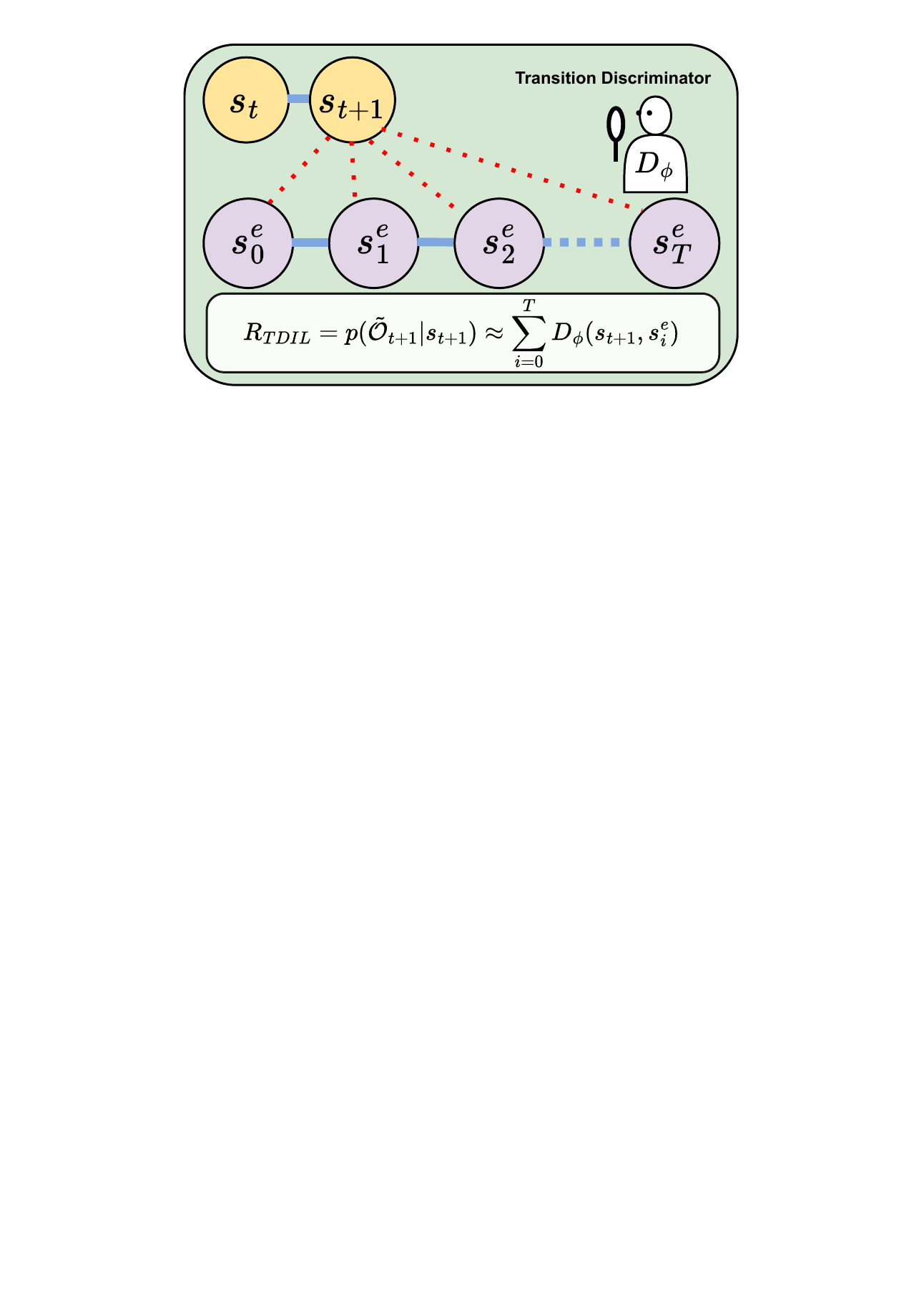}
    \vspace{-1.7em}
    \caption{The approximation of $\Rsur$ through the use of the transition discriminator $D_\phi$.
    }
    \label{fig:transition_discriminator}
	\end{minipage}\hfill
	\begin{minipage}{0.58\linewidth}
	\centering
    \footnotesize
    \includegraphics[width=\linewidth]{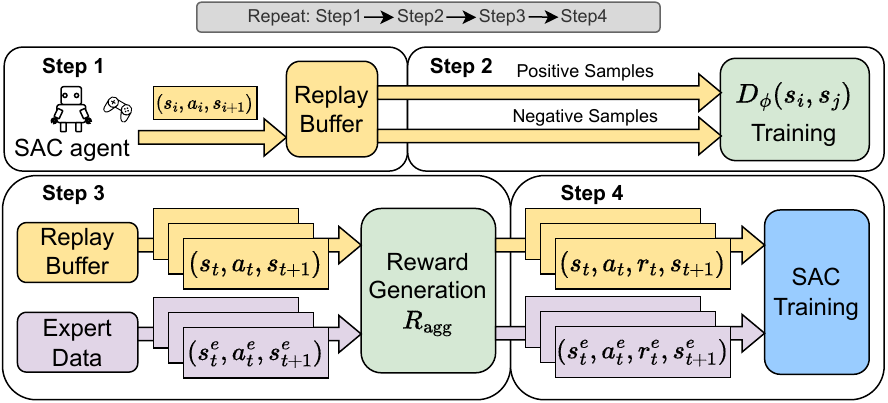}
    \vspace{-2em}
    \caption{
    An overview of the TDIL method. \textbf{Step 1:} Agent-environment interaction. \textbf{Step 2:} Transition discriminator updates. \textbf{Step 3:} Generation of aggregated rewards. \textbf{Step 4:} Training an RL agent based on the generated reward signals.}
    \label{fig:overview}
	\end{minipage}
    \vspace{-1.5em}
\end{figure*}

%% file: Sections/4_5_TD_selection.tex
\subsection{Distinctions of State and State-Action based TDs}
\label{subsec::TD_selection}
While a state-action discriminator could potentially increase the density of reward signals, our TDIL reward is designed based on a state discriminator for the following two reasons. First, the state-based discriminator provides rewards on a larger number of state-action pairs, which can result in denser rewards compared to a state-action based discriminator. For a transition $(s_t, a_t, s_{t+1})$, the state discriminator provides rewards as long as $s_{t+1}$ is in the expert proximity. In contrast, a state-action based discriminator would provide rewards only when $s_t$ is in the expert proximity. Furthermore, training a state-action discriminator $D'(s_i, a_i, s_j)$ can be more challenging since it not only requires ensuring that $s_t$ can transition to $s_j$ but also necessitates validating whether $a_i$ is the permissible action for such a transition.

%% file: Sections/4_6_relative_reward.tex
\subsection{Relative Rewards for Blind Model Selection}
\label{subsec::relative_rewards_for_blind_model_selection}

In practice, we find that the normalized surrogate rewards can effectively select a decent model from a collection of training checkpoints, without the need for direct access to ground truth rewards $R$. This attribute is noteworthy in the context of IL applications, where the best model may not be the one trained for the longest, as detailed in Appendix~\ref{sub::mujoco_relative}. This capability is realized through the computation of relative total rewards $\sum^{T}_{t=0}R_{\text{TDIL}}(s_t,a_t) / \sum^N_{i=0}R_{\text{TDIL}}(s^e_i,a^e_i)$, instead of using the raw return $\sum^{T}_{t=0}R_{\text{TDIL}}(s_t,a_t)$. These relative total rewards serve as a decent indicator for selecting the best-performing model. Note that the implementation details of the relative returns are provided in Appendix~\ref{subsec::blind_model_selection_method}.

%% file: Sections/5_experiments.tex
\section{Experimental Results}
\label{sec::experiments}

\input{Figures/main_experiment}
\input{Figures/correlation}

This section presents our experimental results conducted in two distinct environments: MuJoCo~\citep{todorov2012mujoco} and Adroit Hand~\citep{rajeswaran2017learning}. We also include ablation studies to provide deeper insights into our method.

\label{sec:experiments}
\subsection{Baselines}
\label{sub::exp_setup}

We have selected BC~\citep{bain1995framework}, GAIL~\citep{ho2016generative}, f-IRL~\citep{firl2020corl}, PWIL~\citep{dadashi2021primal}, and CFIL~\citep{freund2023coupled} as our baseline methods. GAIL is a widely recognized and extensively adopted IL method. CFIL and f-IRL both represent the state-of-the-art (SOTA) adversarial-based methods, while PWIL serves as a representative of non-adversarial approaches. IQ-Learn~\citep{garg2021iqlearn} is excluded from our comparison due to its subpar performance~\cite {zeng2022maximumlikelihood, sikchi2022ranking}.  The original f-IRL study did not include evaluations on \textit{Humanoid-v3}. Our assessments revealed that its performance on \textit{Humanoid-v3} was considerably below acceptable levels. LS-IQ~\citep{al-hafez2023lsiq} is also excluded due to the unavailability of its complete code. We do not include AIRL~\cite{fu2017learning} since the previous study~\citep{firl2020corl} has demonstrated that AIRL exhibits inferior performance compared to f-IRL~\citep{firl2020corl}, particularly under settings with few expert demonstrations.

\subsection{Single-Demo IL Evaluation Results and Insights}
\label{sub::mujoco_oracle_select}

In this section, we compare two variants of TDIL described in Section~\ref{subsec::tdil}, with baseline methods under the single-demo IL setting. The first variant, denoted as Ours~($\Rsur$ + BC), replaces $\Rirl$ with BC loss. In this variant, we set $\beta=0$ in $\Rtot$ to assess whether our surrogate reward $\Rsur$ alone can effectively guide the agent back to expert states. Given that $\Rirl$ is not utilized in this variant, BC is applied to ensure that the policy learns the expert demonstration. The second variant, denoted as Ours~($\Rtot$), explores the potential of training SAC agents by $\Rtot$, without the help of BC loss.

We first conduct experiments in five MuJoCo environments, including \textit{HalfCheetah-v3}, \textit{Hopper-v3}, \textit{Ant-v3}, \textit{Humanoid-v3}, and \textit{Walker2d-v3}. All algorithms are trained over 3M timesteps, each employing five random seeds. The expert demonstrations are generated by a well-trained SAC~\citep{haarnoja2018soft} agent using the default parameters. Note that the expert demonstrations we collected have higher total returns than those used by some of the baseline papers, which might lead to different results. For the aggregate reward $\Rtot$ version, we select $\beta=0.9$ based on our grid search, with the results presented in Table~\ref{tab:beta}.

Fig.~\ref{fig:main_experiment} presents the evaluation results on MuJoCo, with the bars representing the normalized performance of each algorithm compared to the expert's performance. Table~\ref{tab:main_table} presents the detailed testing results.
Both versions of our method, i.e., Ours ($\Rsur$ + BC) and Ours ($\Rtot$), achieve expert-level performance across all tested MuJoCo environments. The performance of Ours ($\Rsur$ + BC) indicates that $\Rsur$ can effectively guide agents back to the expert states, even without the use of $\Rirl$. Meanwhile, Ours ($\Rtot$) demonstrates that our proposed reward function, $\Rtot$, effectively serves as a reward mechanism for training SAC agents. In addition to our methods, we evaluated and compared our variants with the baselines. BC does not achieve expert-level performance in any environment, due to overfitting and its inability to generalize to out-of-distribution states. GAIL does not reach expert-level performance either, potentially due to adversarial training instability, sparse rewards in the Maximum Entropy IRL framework in single-demo IL settings, and the algorithm's low sampling efficiency. f-IRL underperforms across various tasks and is particularly ineffective on \textit{Humanoid-v3}. PWIL does not reach expert-level performance, possibly because its reliance on Euclidean distance fails to capture the environmental dynamics correctly. This issue arises especially when two state-action pairs are close in Euclidean distance but unreachable in the MDP, which leads to inaccuracies in the computation of primal Wasserstein cost, as discussed in Section~\ref{sec:motivation}. Lastly, CFIL achieves expert-level performance only on \textit{Hopper-v3}; nevertheless, it shows limited adaptability on \textit{Walker2d-v3}.

\input{Figures/adroitexample}

Besides MuJoCo, we also evaluate TDIL in the Adroit Hand Door environment, as depicted in Fig.~\ref{fig:adroitexample}.
The experimental settings and the complete results are detailed in Appendix~\ref{apx:experiments:adroit}. 
According to the results, while two of the top baselines f-IRL and CFIL show success rates within $40\%$ (mostly $0\%$), both versions of TDIL (i.e., Ours ($\Rsur$ + BC) and Ours ($\Rtot$)) achieve $100\%$ success rates. This demonstrates the generalizability and robustness of TDIL.

\subsection{Blind Model Selection}

To validate the concept discussed in Section~\ref{subsec::relative_rewards_for_blind_model_selection} that relative return can be used in blind model selection,
we graph the ground truth return, raw return, as well as relative return obtained by the agent in an episode during training in Fig.~\ref{fig:correlation}. These rewards are all normalized by their respective maximum values over 3M timesteps. From these plots, a clear positive correlation between the relative return and the ground truth return can be observed across all environments, whereas the trends of the raw return obtained by the agents do not consistently align with the ground truth return. In particular, in \textit{Hopper-v3}, \textit{Ant-v3}, and \textit{Walker2d-v3}, the raw return exhibit a trend that initially rises and then falls over 3M timesteps, a pattern that does not mirror the ground truth rewards. This fluctuation suggests that the accuracy of the transition discriminator grows during training, as it is trained on more data near the expert support collected by the increasingly proficient RL agent. Additional results on blind model selection are available in Appendix~\ref{sub::mujoco_relative}.

\subsection{Ablation Studies}
\label{sub::ablation}

This section examines the effect or performance of several components within TDIL in the Mujoco environments. 

\textbf{Different $\beta$ in $\Rtot$.}
$\beta$ represents the weight of $\Rirl$ in $\Rtot$. The detailed results are reported in Table~\ref{tab:beta}, which reveals that by setting $\beta$ within the range of [0.1, 0.9], the agent consistently achieves expert-level performance without the use of BC loss. This highlights the adaptability and robustness of TDIL, even when certain components, such as $\beta$, are not fine-tuned for specific experimental contexts.

\textbf{Training with Pure $\Rsur$.} 
To evaluate the performance of using $\Rsur$ solely, we present the experimental results on MuJoCo in the ``w/o BC" column in Table~\ref{tab:ablation}. This version exhibits performance inferior to both TDIL variants: Ours ($\Rsur$ + BC) and Ours ($\Rtot$). The decrease in performance highlights the significance of learning expert actions on expert states, which can be realized through the use of $\Rirl$ or BC loss. However, it still achieves comparable performance against all the baselines. For instance, it attains expert-level performance in \textit{Walker2d-v3}, where all the baselines fail.

\textbf{The accuracy of the transition discriminator.}
To ensure the transition discriminators employed are well-trained, we report their accuracy in Table~\ref{tab:tdacc_short}. The experimental results indicate that the transition discriminators can achieve an accuracy of 0.988 or higher across all environments, regardless of the dataset type ($\positiveset$, $\easynegativeset$, or $\hardnegativeset$). This demonstrates that our rewards are derived from well-trained transition discriminators, which can serve as trustworthy approximations employed in Eq.~(\ref{eq:approx_r_tdil}). Moreover, we evaluate the accuracy of the transition discriminators when trained with different $\alpha$ values in Eq.~(\ref{eq:bce}). The results in Table~\ref{tab:tdacc} reveal that the accuracy of a transition discriminator is not sensitive to the selection of hyperparameter $\alpha$. This demonstrates the robustness of our proposed methodology.
\input{Table/ablation}
\input{Table/tdacc_short}

\textbf{Training without $\hardnegativeset$.} 
The column labeled ``w/o $\hardnegativeset$'' in Table~\ref{tab:ablation} illustrates the impact of excluding $\hardnegativeset$ during training. This configuration yields comparable performance across all environments, with the exception of \textit{Humanoid-v3}. This finding implies that in less complex environments, the information contained in hard negative samples may not be crucial. However, in the \textit{Humanoid-v3} environment, the absence of hard negative samples adversely affects performance. This discrepancy may be attributed to the vast state space of \textit{Humanoid-v3}, which diminishes the likelihood that easy negative samples encapsulate the essential information contained in hard negative ones. Furthermore, as \textit{Humanoid-v3} is a more complicated environment, the agent might be sensitive to inaccurately estimated rewards resulting from the absence of hard negative samples.
\paragraph{Multiple expert demonstrations.} Our methodology is not limited to a single demonstration setting. To validate this, we conduct additional experiments with multiple demonstrations and present the results in Table~\ref{tab:multidemo}. These experimental results demonstrate that our method can achieve expert-level performance with additional demonstrations.

%% file: Figures/main_experiment.tex
\begin{figure*}[ht!]
\includegraphics[width=\textwidth]{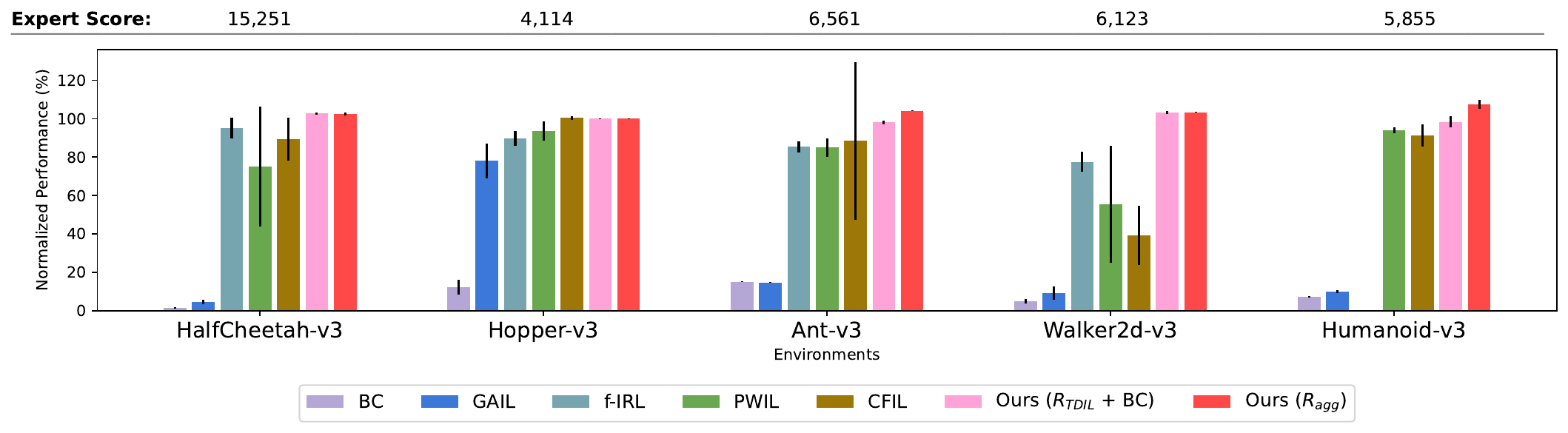}
\centering
\vspace{-2.5em}
\caption{
Normalized performance evaluation of different methodologies using the Oracle model selection under the single-demo setting.
}
\label{fig:main_experiment}
\end{figure*}

%% file: Figures/correlation.tex
\begin{figure*}[ht!]
\vspace{-0.8em}
\includegraphics[width=\textwidth]{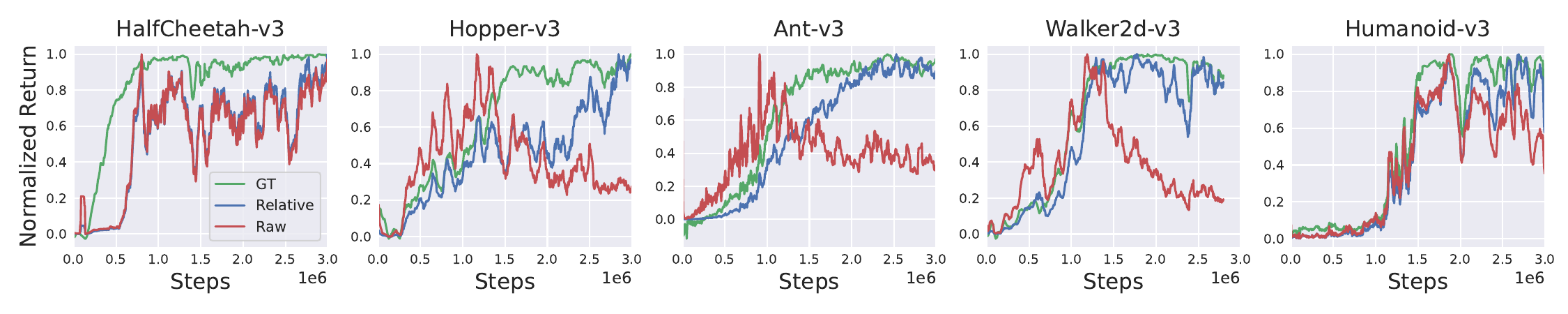}
\centering
\vspace{-2.5em}
\caption{
Comparison of normalized ground truth, raw, and relative return for blind model selection. 
}
\vspace{-1.5em}
\label{fig:correlation}
\end{figure*}

%% file: Figures/adroitexample.tex
\begin{wrapfigure}{r}{0.22\textwidth}
\vspace{-1em}
  \centering
  \includegraphics[width=0.19\textwidth]{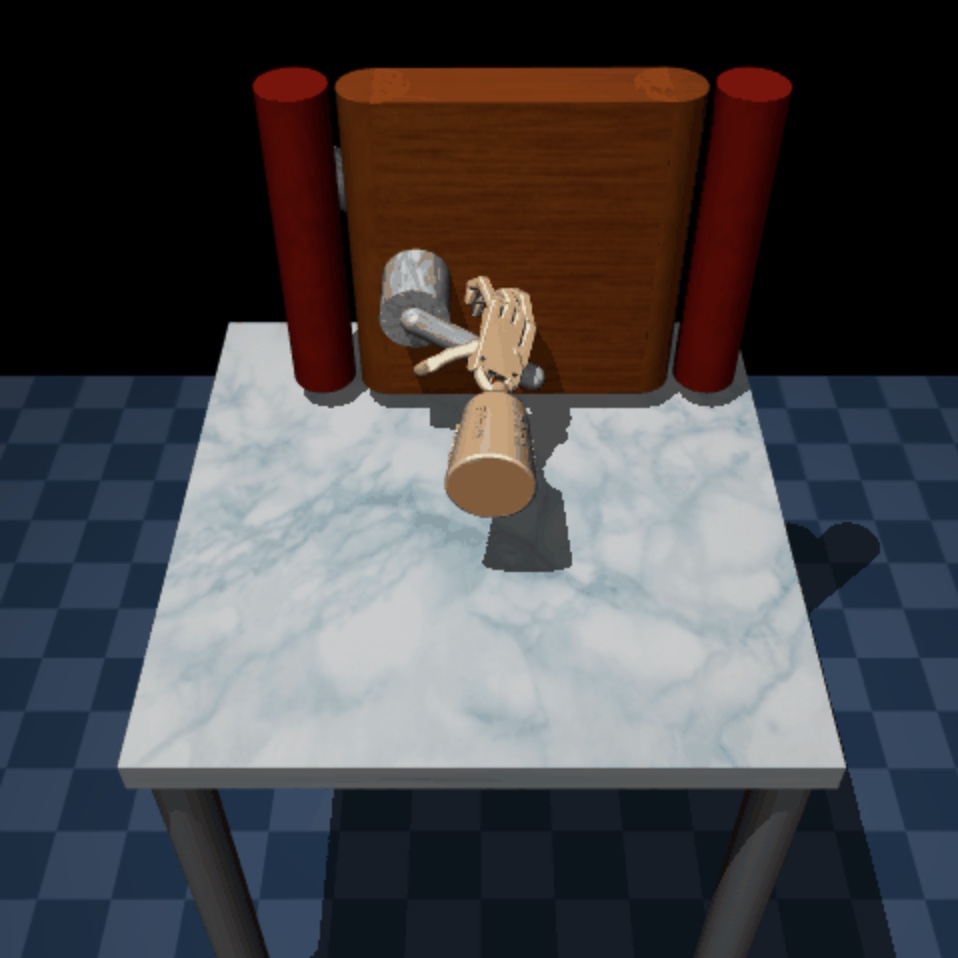}
  \vspace{-0.5em}
  \caption{The Adroit Hand Door environment illustration.}
  \label{fig:adroitexample}
  \vspace{-1em}
\end{wrapfigure}

%% file: Table/ablation.tex
\begin{table}[t]
\renewcommand{\arraystretch}{1.3}
\newcommand{\boldtoprule}{\midrule[1.5pt]}
\scriptsize
\centering
\caption{
TDIL w/ and w/o the BC loss and hard negative samples.}
\vspace{0.5em}
\label{tab:ablation}
\begin{tabularx}{0.45\textwidth}{@{\hspace{1em}}Xcccc@{}}
\boldtoprule
& $\Rsur$ + BC & w/o BC & w/o $\hardnegativeset$ \\ \hline
HalfCheetah-v3 & 15,666 $\pm$ 85 & 12,630 $\pm$ 6,854 & 15,718 $\pm$ 179 \\
Hopper-v3 & 4,115 $\pm$ 14 & 3,890 $\pm$ 562 & 4,143 $\pm$ 6 \\
Ant-v3 & 6,434 $\pm$ 66 & 3,995 $\pm$ 2,408 & 6,571 $\pm$ 116 \\
Humanoid-v3 & 5,758 $\pm$ 173 & 5,575$\pm$196 & 4,868$\pm$2,443 \\
Walker2d-v3 & 6,312 $\pm$ 47 & 6,281 $\pm$ 76 & 6,268 $\pm$ 53 \\ \boldtoprule
\end{tabularx}
\vspace{-2em}
\end{table}

%% file: Table/tdacc_short.tex
\begin{table}[t]
    \renewcommand{\arraystretch}{1.3}
    \newcommand{\boldtoprule}{\midrule[1.5pt]}
    \scriptsize
    \centering
    \caption{Accuracy of the transition discriminators.}
\vspace{0.5em}
    \label{tab:tdacc_short}
    \begin{tabularx}{0.45\textwidth}{@{\hspace{2em}}Xc@{\hspace{3em}}c@{\hspace{3em}}c@{\hspace{2em}}}
    \boldtoprule
        & $\positiveset$ & $\easynegativeset$  & $\hardnegativeset$ \\ \midrule
        HalfCheetah-v3 &  1.0 & 0.992 & 0.996  \\
        Hopper-v3 &  1.0  & 0.996 & 0.996 \\
        Ant-v3 & 1.0  & 0.992 & 0.992 \\
        Humanoid-v3 &  1.0  & 0.99 & 0.988  \\
        Walker2d-v3 & 1.0  & 0.992 & 0.988  \\ \boldtoprule
    \end{tabularx}
    \vspace{-2em}
\end{table}

%% file: Sections/6_related_works.tex
\section{Related Work}
\label{sec:related_work}

\textbf{IL with adversarial training.} Distribution matching methods with a min-max formulation~\citep{ho2016generative, fu2018learning, ke2021imitation, pmlr-v100-ghasemipour20a, firl2020corl, pmlr-v139-swamy21a, Kostrikov2020Imitation, camacho2021sparsedice, freund2023coupled, kostrikov2018discriminatoractorcritic, han2022robust, zeng2022maximumlikelihood, viano2022proximal} might induce potential instability and sub-optimality in situations with sparse demonstration data, which could compromise the effectiveness and reliability of these methodologies.

\textbf{IL with support estimation.} Methods that rely on expert support estimation~\citep{pmlr-v97-wang19d, Brantley2020Disagreement-Regularized, liu2020energy, kim2020imitation} often face difficulties when expert data are limited. This is attributable to their reliance on the availability and quality of expert demonstrations, leaving them ill-suited for scenarios with scarce expert data. 

\textbf{IL with optimal transport.} IL approaches that utilize optimal transport technique~\citep{dadashi2021primal, xiao2019wasserstein}, on the other hand, are also less suitable for the single-demonstration IL setting, as they tend to overlook environmental dynamics. Specifically, these approaches might identify certain states as being close or similar based on their geometric distances of the state space or some state representations~\citep{haldar2023watch}, even though these states may not be permissible for transition in a Markov Decision Process (MDP). This limitation impairs their capacity to capture the complexity and variability of environments. 

\textbf{IL with meta-demonstrations.}
In \textit{one-shot IL}~\citep{duan2017one, finn2017one, yu2018one, pmlr-v155-dasari21a, yu2018one_, 9812450, 8967761, pmlr-v162-netanyahu22a, valassakis2022demonstrate, hu2020two}, researchers have explored the use of meta-demonstrations, which are demonstrations associated with other tasks, as a tool for pre-training before proceeding to one-shot adaptation. However, gathering a substantial volume of meta-demonstrations, which are necessary for training meta parameters prior to their one-shot utilization, can be infeasible due to the expensive nature of expert demonstrations. Note that these studies are orthogonal to our work.

\textbf{IL with Ground Truth Reward Function.} Some previous studies~\citep{aytar2018playing, wu2021learning, peng2018deepmimic} focus on improving the RL agent with the help of a single demonstration. However, they still allow their RL agent to access the ground truth reward function of the environment.

The majority of the aforementioned methods either struggle to achieve expert-level performance in high-dimensional environments or are less adept at achieving robust generalization~\citep{firl2020corl, freund2023coupled, dadashi2021primal, al-hafez2023lsiq}. These constraints highlight the necessity for enhanced strategies in the single-demonstration IL setting. The key objectives include accommodating limited expert data while taking into account environmental dynamics. Please refer to Appendix~\ref{apx:extended_related_work} for more details.

%% file: Sections/7_conclusion.tex
\section{Conclusions and Future Works}
\label{sec:conclusion}
In this paper, we proposed TDIL as a robust approach to address the challenges inherent in single-demo IL settings. By considering the transitions towards expert states as reasonable, we defined a dense surrogate reward function that can be approximated by a transition discriminator. Our experiments on the MuJoCo benchmarks and the Adroit Hand Door task revealed that our method consistently achieves expert-level performance and outperforms all the baseline algorithms, including BC, GAIL, f-IRL, PWIL, and CFIL. To further validate our surrogate reward function, we compared the ground truth return, raw return, and relative return, and revealed a strong correlation among them. This correlation substantiates the efficacy of our surrogate reward function for blind model selection. Furthermore, we conducted a series of ablation studies to validate the design choices behind TDIL. This work not only provides valuable insights but also lays a solid groundwork for future exploration in single-demo IL settings.

To accommodate more complex or higher-dimensional environments, a promising future direction involves extending our surrogate rewards from one-step to multi-step transitions, as briefly described in Appendix~\ref{apx:multi_step_expert_proximity}. This would enable the surrogate reward function to provide rewards for a broader range of transitions and guide the agent back to the expert state more efficiently.

%% file: Sections/8_acknowledgement.tex
\section*{Acknowledgment}
\label{sec:acknowledgement}
The authors gratefully acknowledge the support from the National Science and Technology Council (NSTC) in Taiwan under grant number MOST 111-2223-E-007-004-MY3, the financial support from MediaTek Inc., Taiwan, and the support from the National Science Foundation (NSF) under grant numbers 2048280, 2331966, 2325121, and 2244760, as well as from the Office of Naval Research (ONR) under grant number N00014-23-1-2300:P00001.
The authors would also like to express their appreciation for the donation of the GPUs from NVIDIA Corporation and NVIDIA AI Technology Center (NVAITC) used in this study. Moreover, the authors extend their gratitude to the National Center for High-Performance Computing (NCHC) in Taiwan for providing the necessary computational and storage resources.

%% file: Sections/9_impact_statement.tex
\section*{Impact Statement}
This research focuses on single-demo IL and presents potential advancements for robotic and RL agent training. It aims to improve the training stability and efficiency for RL agents. It relies on publicly available data/environments, which ensures transparency and avoids the use of sensitive or proprietary information. Our method eliminates the complexity and time required for designing delicate rewards and collecting a tremendous amount of expert data. This elimination leads to more intuitive and efficient robot deployment. Specifically, our method enables the learning of complex tasks using only a single expert demonstration and can potentially enhance productivity and adaptability. Our approach has the potential to positively impact society by making robotic systems more accessible and efficient.

%% file: Sections/all_appendix.tex
\input{Sections/a0_appendix.tex}
\input{Sections/a1_extended_related_work}
\input{Sections/a2_algorithm_and_training_detail}
\input{Sections/a3_configuration}

\input{Sections/a4_experiments}
\input{Sections/a5_multi_step_expert_proximity}

%% file: Sections/a0_appendix.tex

\setcounter{section}{0}
\setcounter{equation}{0}
\setcounter{figure}{0}
\setcounter{table}{0}

\renewcommand{\thefigure}{A\arabic{figure}}
\renewcommand{\thetable}{A\arabic{table}}
\renewcommand{\theequation}{A\arabic{equation}}

\section{Appendix}
\label{apx}
In this appendix, we provide review of related works, detailed training configurations, additional experimental results, and discussions on the proposed TDIL method. In Section~\ref{apx:extended_related_work}, a detailed review of previous works is provided. In Section~\ref{apx:algo_detail}, we provide the training detail of the proposed TDIL algorithm. In Section~\ref{apx:configuration}, we elaborate on the experimental setups as well as the model architecture adopted in our method. In Section~\ref{apx:experiments}, we present additional experimental results to validate the effectiveness of our method.  In Section~\ref{apx:multi_step_expert_proximity}, we extend the expert reachability indicator $\K_t$ to multiple timesteps.

%% file: Sections/a1_extended_related_work.tex
\subsection{Extended Review of Related Work}
\label{apx:extended_related_work}

The single-demonstration IL setting presents a unique and challenging problem domain. Earlier online IL research commonly treats this setting as a component of their ablation studies, and often overlooks its significance. This section engages in a discussion about several well-known online IL algorithms, which can be broadly grouped into two categories: adversarial-based and non-adversarial-based methods. This discussion allows us to highlight our differences from these prior methods, and delve into the priminary reasons behind their limited effectiveness in the single-demonstration IL context.

\textbf{Adversarial-based methods:} Adversarial-based approaches aim to align the agent's state, state-action, or state-next-state distributions with those of the expert by employing various divergence or distance measures. For instance, GAIL~\citep{ho2016generative} adopts the GAN-like framework~\citep{goodfellow2020generative} to train a discriminator and minimize Jensen-Shannon divergence. AIRL~\citep{fu2018learning}, on the other hand, utilizes forward KL-divergence to derive stationary rewards and enhance transfer learning.  Building upon these approaches, the authors of~\cite{ke2021imitation} and f-MAX~\citep{pmlr-v100-ghasemipour20a} unify GAIL and AIRL under the umbrella of f-divergence.  To further extend these methods, f-IRL~\citep{firl2020corl} uses gradient descent to recover a stationary reward function from the expert density.  In addition, recent research by the authors of~\cite{pmlr-v139-swamy21a} suggest that various forms of IL can be understood as moment matching under different assumptions. Another line of work is based on the DICE~\citep{nachum2019dualdice} framework.  For example, ValueDICE~\citep{Kostrikov2020Imitation} utilizes the Donsker-Varadhan formulation of KL-divergence to develop an off-policy method, while SparseDICE~\citep{camacho2021sparsedice} introduces a regularizer to enable training with sparse expert data.  Inspired by ValueDICE, CFIL~\citep{freund2023coupled} trains a pair of normalizing flows to optimize the Donsker-Varadhan representation of KL-divergence.  Moreover, a variety of research efforts~\citep{kostrikov2018discriminatoractorcritic, han2022robust, zeng2022maximumlikelihood, viano2022proximal} have been directed towards addressing specific challenges within the field. For instance, DAC~\citep{kostrikov2018discriminatoractorcritic} modifies GAIL to facilitate off-policy training and concurrently tackles reward bias issues. MD-AIRL~\citep{han2022robust} enhances robustness by incorporating mirror-descent into AIRL. In a further effort to improve efficiency, both ML-IRL~\citep{zeng2022maximumlikelihood} and $\text{P}^2\text{IL}$~\citep{viano2022proximal} have been designed to relax the nested policy evaluation and cost optimization loop.  Most of the above methods, while being successful in online IL, do not perform well in the single-demonstration IL setting. The reason behind this can be attributed to two primary factors. The first factor is that the majority of their objectives typically align with a min-max formulation, which could lead to unstable training, especially in situations with limited data. The second factor is inherent to their distribution-matching nature, which necessitates taking expectations over the expert distribution. Nevertheless, this process could become unreliable when dealing with sparse expert data.  In contrast to these previous approaches, our methodology does not seek to match the distribution of the agent with that of the expert. This different approach avoids the issues of inaccurate expectations and unstable adversarial training.

\textbf{Non-adversarial based method:} Non-adversarial based methods often aim to circumvent unstable training by designating stationary rewards to guide the agent toward expert behavior. Examples include SQIL~\citep{reddy2019sqil}, D2-Imitation~\citep{sun2022deterministic}, and ILR~\citep{ciosek2022imitation}, which implement a binary reward scheme that assigns a value of $1$ to expert data and $0$ to agent data. These methods typically require a substantial amount of expert data to achieve optimal performance in practice. Another line of research explores a two-stage training approach, wherein a reward surrogate is first trained offline and then utilized during interaction with the environment. For instance, RED~\citep{pmlr-v97-wang19d} estimates expert support by leveraging Random Network Distillation~\citep{burda2018exploration}, while DRIL~\citep{Brantley2020Disagreement-Regularized} pretrains an ensemble of Behavior Cloning (BC)~\citep{pomerleau1991efficient} models and employs their variance as a cost function. EBIL~\citep{liu2020energy} and NDI~\citep{kim2020imitation} employ density models, such as Energy-Based Models (EBM)~\citep{song2021train} and Masked Autoencoder Density Estimation (MADE)~\citep{germain2015made}, to estimate expert support density.  Nevertheless, these methods necessitate a significant amount of expert data for training the offline reward surrogate, which poses challenges when applied to the single-demonstration setting. Another non-adversarial approach, PWIL~\citep{dadashi2021primal}, attempts to minimize discrepancy between an agent's and an expert's distributions by employing the primal form of Wasserstein distance. This method requires the computation of the Euclidean distance between every state-action pair and those of the expert, a measure that may not precisely align with the distance as defined by the Markov Decision Process (MDP). In contrast, our method
takes the properties of the underlying MDP into account. Furthermore, recent advancements such as IQ-Learn~\citep{garg2021iqlearn} and LS-IQ~\citep{al-hafez2023lsiq} offer a unique perspective, as they implicitly represent policy and reward using a single Q-function. 
Nevertheless, according to our experiments, these methods could suffer from instability during training and may not consistently perform well across various IL tasks.

%% file: Sections/a2_algorithm_and_training_detail.tex
\subsection{Algorithm and Training Details}
\label{apx:algo_detail}
\subsubsection{Practical Algorithm}
The training process concurrently updates the transition discriminator and the SAC agent. Both the agent and expert transition data are utilized to train the SAC agent, with the agent's reward calculated using the transition discriminator. The reward calculation method involves the computation of the reward of both agent data and expert data. The agent reward is calculated by pairing the next state $s_{t+1}$ of a transition $(s_t, a_t, s_{t+1})$ with ``every'' expert state from the demonstration, as illustrated in Fig.~\ref{fig:transition_discriminator}, and using the transition discriminator to calculate the reachability probability of each pair. These probabilities are then summed to yield a reward $r_t=\sum^T_{i=0}D_\phi(s^a_{t+1},s^e_i)$. The expert rewards are computed in a similar manner by pairing each next state of an expert transition with every other expert state, and summing the resulting probabilities.
\subsubsection{Training stabilization}
To ensure stable training, a target transition discriminator, denoted as $\hat{D}$, is employed in our training process to compute the reward. $\hat{D}$ is soft-updated using the formula $\hat{D}=(1-\lambda)D+\lambda \hat{D}$, where $\lambda$ is a hyperparameter set to 0.0001 in practice. The target transition discriminator helps mitigate the instability caused by SGD training, providing a more stable and consistent target for the SAC agent to learn from. This reduces overfitting and other potential sources of instability, making the training process less susceptible to fluctuations and ensuring a consistent trajectory towards convergence.
\subsubsection{Algorithm detail}
Algorithm~\ref{algo:pseudo} presents a practical training methodology of the proposed method, refering to TDIL. It takes as input the policy $\pi$ of an imitator agent, an environment $\mathcal{E}$, a replay buffer $B$, a Transition Discriminator $D$, a Target Transition Discriminator $\hat{D}$, and an expert trajectory $\tau^e$. The output is a trained optimal agent $\pi^*$.
The training process is iterative, continuing until a convergence criterion is met. During each iteration, the policy $\pi$ interacts with $\mathcal{E}$, and the states and actions $(s_t, a_t, s_{t+1})$ are stored in $B$.
Next, $D$ is updated based on Eq.~(\ref{eq:bce}) based on the stored transitions. Following this, $\hat{D}$ is soft-updated by $D$, which help stabilizing training.
The algorithm then samples a batch of transitions from both $B$ and the expert trajectory $\tau^e$, and calculates the reward using $\hat{D}$. This reward is then used to update $\pi$ by comparing the agent's transitions with those of the expert.
Finally, $\pi$ is updated using a BC loss, denoted as $L_{BC}=\text{MSE}(a\sim\pi(s_i^e),a_i^e)$, which aims to minimize the discrepancy between the agent's actions and the expert actions. Through the repetition of these steps, the TDIL algorithm trains the imitator agent $\pi$ to match the expert's performance in the given environment.
To satisfy the policy assumption in Section~\ref{sec:background}, the BC loss $L_{BC}$ is included to ensure $p(\O_t=1|s_t)=\max_a p(\O_t=1|s_t,a)$.

\input{Algorithms/algo1v2}
\input{Figures/toyexpertproximity}

%% file: Algorithms/algo1v2.tex
\RestyleAlgo{ruled}
\begin{algorithm}[ht]
    \SetAlgoLined
    \LinesNumbered
    \caption{TDIL: IL via Transition Discriminator}
    \label{algo:pseudo}
    \SetKwInOut{Input}{Input}
    \SetKwInOut{Output}{Output}
    
    \Input{Imitator Agent $\pi$, Environment $\mathcal{E}$, Replay Buffer $B$, Transition Discriminator $D$, Target Transition Discriminator $\hat{D}$, Expert Trajectory $\tau^e$}
    \Output{Trained optimal agent $\pi^*$}
    \While{not converge}{
        $\pi$ interacts with $\mathcal{E}$, storing $s_t, a_t, s_{t+1}$ in $B$
        
        Update $D$ with Eq.~(\ref{eq:bce})
        
        Soft-update $\hat{D}$ with $D$
        
        Sample one batch of transition from $B$ and $\tau^e$, and calculate the reward with $\hat{D}$
        
        Update $\pi$ using sampled agent transitions and expert transitions with calculated reward
        
        Update $\pi$ with $L_{BC}$
    }
\end{algorithm}

%% file: Figures/toyexpertproximity.tex
\begin{figure}
  \centering
  \hspace*{7em}
  \includegraphics[width=0.5\textwidth]{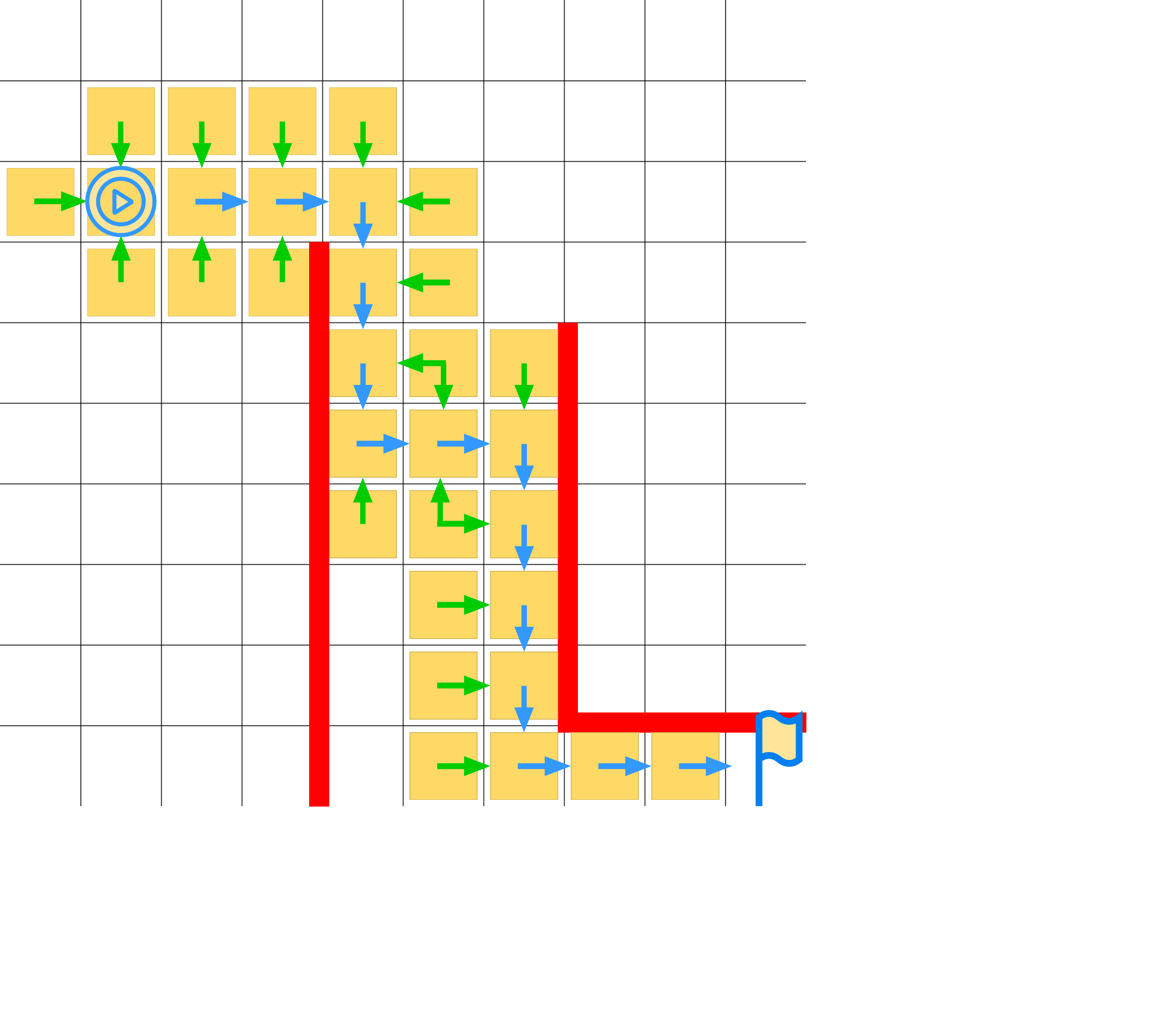}
  \vspace{-5em}
  \caption{Visualization of the relationship between expert demonstration and expert proximity in the grid-world}
  \label{fig:expertprox}
\end{figure}

%% file: Sections/a3_configuration.tex
\subsection{Experimental Setups}
\label{apx:configuration}
\subsubsection{Model Architecture of TDIL}
In this section, we provide the implementation details of TDIL. The backbone of TDIL is built upon the Soft Actor-Critic (SAC) framework. The actor and critic networks in SAC are implemented as neural networks with three hidden layers and rectified linear unit (ReLU) activation functions. Each of these hidden layers consists of 256 nodes. 
The actor’s output is first projected to $[-1,1]$ using a hyperbolic tangent (tanh) function, and then scaled to  the value range required by the environments.
\subsubsection{Implementation of $\Rirl$}
The implementation of $\Rirl$ leverages the discriminator in GAIL. This demonstrates that, even when paired with a basic IRL reward function, the proposed $\Rsur$ can effectively guides the agent toward expert proximity and learn the expert behavior, leading to expert-level performance. 
\subsubsection{Code Implementation and Hardware Configuration}
The code implementation and expert data used in this work are available on this \href{https://github.com/stanl1y/tdil}{GitHub repository}. The computational requirements for the experiments presented in Section~\ref{sec:experiments} is elaborated in Table~\ref{tab:machine}.
\input{Table/machine}

%% file: Table/machine.tex
\begin{table*}[ht]
\renewcommand{\arraystretch}{1.4}
\newcommand{\boldtoprule}{\midrule[1.2pt]}
\centering
\footnotesize
\caption{The hardware specification used to perform our experiments.}
\label{tab:machine}
\begin{tabularx}{\textwidth}{l|>{\centering\arraybackslash}X}
    \boldtoprule
    Hardware & Specification \\
    \boldtoprule
    RAM & 128GB \\
    CPU & AMD Ryzen Threadripper 3990X 64-Core Processor \\
    GPU & NVIDIA GeForce RTX 3090 \\
    \boldtoprule
\end{tabularx}
\vspace{-0.5em}
\end{table*}

%% file: Sections/a4_experiments.tex
\subsection{Additional Experiments}
\label{apx:experiments}
In this section, we provide additional experimental results and discussions. In Section~\ref{apx:experiments:training_curve}, we present the training curves of the proposed and the baseline methods to demonstrate the performance and stability of different algorithms. In Section~\ref{apx:experiments:baseline_with_bc}, we compare TDIL with baselines trained with additional BC loss. In Section~\ref{apx:experiments:adroit} we present the success rate curve of the proposed and the baseline method in Adroit Hand environment. In Section~\ref{subsec::blind_model_selection_method}, we give detailed analysis on the relative rewards for blind model selection. In Section~\ref{sub::mujoco_relative}, we offer the evaluation results the models selected according to different blind selection metrics during training for demonstrating the effectiveness of the proposed blind selection method. In Section~\ref{apx:experiments:accuracy}, we examine the influences of the hard negative samples on the performance of the transition discriminators under various scenarios. Finally in Section~\ref{apx:experiments:alpha} and Section~\ref{apx:experiments:beta}, we investigate the influence of different choices of the hyper-parameter $\alpha$ and $\beta$ respectively.
\input{Table/main_table}
\input{Figures/training_curve}
\subsubsection{Training Curves}
\label{apx:experiments:training_curve}
Fig.~\ref{fig:training_curve} presents the training curves of TDIL as well as the other baseline methods, including BC, GAIL, f-IRL, PWIL, and CFIL. It is worth noting that the optimization of CFIL in the \textit{HalfCheetah-v3} environment is numerically unstable as its output values sometimes become NaN during the training process. As a result, the training curve of CFIL in the \textit{HalfCheetah-v3} environment can only be plotted partially.
Fig.~\ref{fig:training_curve} demonstrates that TDIL is capable of reaching the expert level and exhibits a consistently stable training process across different environments compared to the other baselines.

\subsubsection{Experiments on Using Multiple Expert Demonstrations}
\label{apx:experiments:multiple_demo}
\input{Table/multidemo}
The TDIL method is designed to address the challenging limitations of the single-demo IL setting. TDIL can be regarded as an approach that leverages all available information from the expert data under the constrained condition of limited expert demonstration. However, it is not restricted to single-demo IL. Although providing more expert demonstrations might be beneficial, it does not significantly affect the performance. This is because TDIL can achieve expert-level performance with only a single expert demonstration, as shown in Table~\ref{tab:multidemo}. Furthermore, if more expert demonstrations are available, it may not be necessary to learn a transition discriminator, and other state-of-the-art IL techniques can be employed. These techniques, however, may not adequately address the single-demo problem that TDIL is specifically designed to tackle.

\subsubsection{Performance comparison between TDIL and baselines with BC loss}
\label{apx:experiments:baseline_with_bc}
\input{Table/baselinebc}
We have conducted additional experiments to provide a more comprehensive analysis on adding BC loss into the training process of baselines. Table~\ref{tab:baselinebc} presents the performance of CFIL, PWIL, and TDIL with BC loss, directly compared with training the agent with $R_{\text{TDIL}}$ and BC loss. Notably, some baselines demonstrate improved performance with BC loss, yet TDIL consistently outperforms all baselines. It is noteworthy that CFIL exhibited a substantial performance boost with BC loss in the Walker2d-v3 environment. However, it is crucial to acknowledge that CFIL encountered numerical issues, specifically the occurrence of actor output becoming NaN in the middle of training across all environments. This highlights potential instability in CFIL algorithm.

\subsubsection{Experiments in Adroit hand environment}
\label{apx:experiments:adroit}

\input{Figures/adroit}
In Fig.~\ref{fig:adroit}, we present the experiment in the AdroitHandDoor environment. 
The AdroitHandDoor environment is a component of the Adroit manipulation platform, featuring a Shadow Dexterous Hand attached to a free arm with up to 30 actuated degrees of freedom~\citep{rajeswaran2017learning}. 
We do not evaluate TDIL on the other adroit tasks since the agent is required to achieve different goals encapsulated within the state feature. In such environments, agents cannot learn the meaning of different goals if only one expert demonstration is offered. This limitation arises because all expert states in the provided single demonstration inherently possess the same goal, which restricts the agent's comprehension of the goal feature. As a result, the agent might not learn the goal feature adequately, and this can result in a policy that fails to condition on the goal effectively. For instance, the agent could mimic an expert trajectory without adapting to changes in the goal.

In the AdroitHandDoor-v1 scenario, the task involves undoing a latch and swinging open a door with a biased torque that keeps it closed. The environment, based on a 28-degree-of-freedom system, includes a 24-degree-of-freedom ShadowHand and a 4-degree-of-freedom arm. The action space is represented as a Box(-1.0, 1.0, (28,), float32), with control actions specifying absolute angular positions of the hand joints. The observation space is a Box(-inf, inf, (39,), float64), containing information on finger joint angles, palm pose, and the state of the latch and door, Fig.~\ref{fig:adroitexample} illustrates the task.

The episode's time step limit is set at 200. During the testing phase, the agent undergoes perturbation through five time-steps of random actions in the beginning of the episode to enhance difficulty and introduce stochasticity. In comparison to BC and two of the top-performing baselines from the main experiment, the results demonstrate that TDIL attains an expert-level performance within 1 million steps, surpassing the performance of BC, PWIL, and CFIL.

\subsubsection{Exploring Relative Rewards for Blind Model Selection}
\label{subsec::blind_model_selection_method}

Blind model selection refers to the process of choosing the optimal model checkpoint throughout the training phase, holds significant importance in the field of IL. In IL, it is generally assumed that obtaining the ground truth reward from the environment is unfeasible, even during testing. This issue, often neglected in prior research, warrants considerable attention. Although the reward signals proposed in this work, denoted as $R_{\text{TDIL}}$, can effectively train the agent, they may not be ideally suited for blind model selection. As training progresses, a potential decrease in the agent's raw rewards is observed. The reduction in raw agent reward may not necessarily signify a decrease in agent's performance; rather, it mirrors the enhanced accuracy of the transition discriminator. As a result, it becomes imperative to establish an indicator that is strongly correlated with the ground truth reward. Such an indicator would facilitate reliable model selection in IL. To meet this requirement, we introduce the concept of `relative reward,' which is denoted as $r_{\text{relative}}$ and is defined as follows:
\begin{equation}
\label{eq:relative_reward}
\begin{aligned}
r_{\mathrm{relative}}=r_{\mathrm{raw\,agent}} / r_{\mathrm{raw\,expert}},
\end{aligned}
\end{equation}
where $r_\mathrm{raw\, agent}=\sum^{\tilde{T}}_{t=0}R_{\text{TDIL}}(s_t,a_t)$ and  $r_\mathrm{raw\, expert}=\sum^T_{t=0}R_{\text{TDIL}}(s^e_t,a^e_t)$ are the total rewards along the agent's and expert's trajectories, and $\tilde{T}$ is the length of the agent's trajectory.
As the transition discriminator may improve its accuracy during training, our aim is to mitigate the influence of its accuracy on reflecting the true extent of reward signals. In an ideal scenario, the reward for expert actions should be higher, while those outside the expert support should be lower.  With this in mind, the essence of Eq.~(\ref{eq:relative_reward}) is to calculate the relative reward by dividing the raw agent reward, derived from the transition discriminator, by the raw expert reward, also derived from the transition discriminator. This process aids in neutralizing the impact of potential inaccuracies of the transition discriminator.  The rationale behind this approach is the presumption that the inaccuracies in the transition discriminator would affect both the raw agent reward and the raw expert reward in a similar fashion.  Hence, when the raw agent reward is divided by the raw expert reward, any inaccuracies that potentially exist in the transition discriminator should theoretically cancel out. This is because these inaccuracies are likely to proportionally affect the numerator (i.e., the raw agent reward) and the denominator (i.e., the raw expert reward) of the division.  For example, if the transition discriminator is consistently underestimating or overestimating the rewards, both the raw agent reward and the raw expert reward would be underestimated or overestimated to a comparable extent.  As a result, their ratio (i.e., the relative reward) should still provide a reliable comparison of agent performance relative to the expert, even if the absolute reward values are incorrect.  This approach, therefore, helps to render the reward calculation more robust to the inaccuracies of the transition discriminator, and enhances the reliability of the model selection process in the single-demonstration IL context.

\input{Figures/blind}
\subsubsection{Blind Model Selection Experiments}

\label{sub::mujoco_relative}
\input{Table/main_blind}
To further substantiate the efficacy of utilizing relative rewards in blind model selection, we performed a MuJoCo experiment in which the optimal testing model was selected without any access to the environmental ground truth rewards. In this experiment, our method used relative rewards as an indicator. In contrast, PWIL employed the Wasserstein distance, following the methodology of the original paper. For the remaining methods, which did not provide an indicator for model selection in their original manuscripts, we chose the model with the lowest policy loss. Table~\ref{tab:main_blind} presents the ratio of performance decrease of each method, which is calculated according to $\frac{blind\ result\ -\ oracle\ result}{oracle\ result}$. The results reveal that our proposed method outperforms both policy loss-based model selection and Wasserstein distance-based model selection schemes. This outcome suggests that relative rewards can effectively guide the selection of the best model, and provides a valuable insight that can be applied in future single-demonstration IL research to develop similar indicators for practical use.

To demonstrate the effectiveness of the blind model selection strategy over the model selection methods adopted by the baselines, we compare the returns obtained using the proposed strategy and the baseline methods along with the highest testing return achieved by each agent during its training process. Fig.~\ref{fig:blind} presents the results of the above setting. In the figure, the blue and red curves represent the total return obtained by each agent and the model selection strategy metric employed by each baseline, respectively. In addition, the solid and the dashed lines depict the highest testing return achieved by each agent during its training process and the return determined by the blind selection strategy, respectively. It is observed that our method is effective in selecting a model with high performance, as the distance between the solid and the dashed lines shown in Fig.~\ref{fig:blind}~(d) is the closest as compared to those depicted in Figs.~\ref{fig:blind}~(a), (b), and (c).  Please note that the returns of the baseline methods can be derived using either the returns of the agent in the last step or the returns selected according to their respective blind selection metrics. In Table~\ref{tab:main_blind} of the main manuscript, we report the higher returns achieved by the baseline methods.

\subsubsection{An Analysis of the Accuracy of the Transition Discriminator}
\label{apx:experiments:accuracy}
\input{Figures/accuracy}

Fig.~\ref{fig:acc_td} illustrates the accuracy of the transition discriminator evaluated on positive samples, easy negative samples, and hard negative samples. Of particular interest is the accuracy of the hard negative samples. In \textit{HalfCheetah-v3} and \textit{Ant-v3} (i.e., Figs.~\ref{fig:acc_td}(a) and (b), respectively), the transition discriminator trained without the use of hard negative samples demonstrates similar accuracy to the one trained with hard negative samples. However, in \textit{Humanoid-v3} (Fig.~\ref{fig:acc_td}(c)), the transition discriminator trained without hard negative samples exhibits significantly lower accuracy compared to the one trained with hard negative samples. These findings substantiate the assumption presented in Section~\ref{sub::ablation}, which suggests that the set of hard negative samples falls within the subset of easy negative samples. In relatively less complex environments, the agent can extract the information embodied in hard negative samples even when training exclusively with easy negative samples. However, this scenario is less probable in the more demanding \textit{Humanoid-v3} environment, leading to the observed discrepancy in accuracy between the two training settings.
These experimental results highlight the importance of incorporating hard negative samples, particularly in complex environments, to improve the accuracy and effectiveness of the proposed transition discriminator.
\subsubsection{Sensitivity analysis on the hyper-parameter $\alpha$ }
\label{apx:experiments:alpha}
\input{Figures/alpha}
\input{Table/tdacc}
We have addressed the sensitivity of the proposed TDIL algorithm to different values of the hyper-parameter $\alpha$ by presenting the corresponding training curves in Fig.~\ref{fig:alpha}. The introduction of the balancing factor $\alpha$ for positive and negative samples aims to mitigate the impact of false negative samples within the pool of easy negative samples. These easy negative samples are composed of pairs of individually randomly sampled states from the replay buffer, and there exists a chance that these pairs may form valid transitions under the Markov Decision Process (MDP), effectively becoming positive samples.

To safeguard the training of the transition discriminator against the adverse effects of false negatives, we assign a smaller weight to negative samples compared to positive samples. Experimental results demonstrate that when $\alpha$ is set to small values (e.g., 0.5, 0.67), the agent takes longer to reach optimal performance in certain environments. Conversely, when the value of $\alpha$ is set to 0.99, the algorithm consistently performs well across various environments. This observation underscores the importance of choosing the hyper-parameter $\alpha$ to ensure optimal and robust performance of the TDIL algorithm.
\subsubsection{Experimental result on different choices of hyper-parameter $\beta$ }
\label{apx:experiments:beta}
\input{Table/beta}
The ablation study on the hyper-parameter $\beta$ is comprehensively presented in Table~\ref{tab:beta}, shedding light on its impact within the overall reward function. By aggregating $R_{\text{TDIL}}$ and $R_{\text{IRL}}$ with a judicious selection of $\beta$, the agent consistently attains expert-level performance guided by this composite reward. Experimental findings suggest that values of $\beta$ within the range of [0.1, 0.9] yield favorable results across various MuJoCo environments. This observation underscores the intrinsic ability and efficacy of the reward function $R_{\text{agg}}$.

Importantly, these results indicate that setting $\beta$ to zero, as done in the main experiments, can still produce effective outcomes, particularly when approximating the effect of $R_{\text{IRL}}$ through BC loss. This pragmatic approach not only maintains computational efficiency but also highlights the adaptability and robustness of the proposed TDIL method, even when certain components, such as $\beta$, are tuned or simplified for specific experimental contexts.

%% file: Table/main_table.tex
\begin{table*}[t]
\renewcommand{\arraystretch}{1.3}
\newcommand{\boldtoprule}{\midrule[1.5pt]}
\centering
\vspace{-0.5em}
\caption{
Performance evaluation of different methodologies using the oracle model selection.
}
\label{tab:main_table}
\vspace{0.25em}
\resizebox{0.95\linewidth}{!}{
\begin{tabular}{ccccccccc}
\boldtoprule
                                    & BC   & GAIL & f-IRL & PWIL  & CFIL & Ours ($\Rsur + $ BC) & Ours ($\Rtot$)  & Expert                             \\ \hline
\multicolumn{1}{l|}{HalfCheetah-v3} & 211 $\pm$ 49 & 693 $\pm$ 158  & 14,560 $\pm$ 823 & 11,460 $\pm$ 4,774 & 13,636 $\pm$ 1,695 & \textbf{15,666} $\pm$ 85& \multicolumn{1}{c|}{15,624 $\pm$ 119} & 15,251 \\ 
\multicolumn{1}{l|}{Hopper-v3}      & 507 $\pm$ 161 & 3,209 $\pm$ 372 & 3,693 $\pm$ 162  & 3,849 $\pm$ 209   & \textbf{4,131} $\pm$ 34 & 4,115 $\pm$ 14  & \multicolumn{1}{c|}{4,115 $\pm$ 14} & 4,114 \\ 
\multicolumn{1}{l|}{Ant-v3}         & 990 $\pm$ 6  & 966 $\pm$ 22  & 5,597 $\pm$ 194  & 5,579 $\pm$ 314   & 5,812 $\pm$ 2,692 & 6,434 $\pm$ 66 & \multicolumn{1}{c|}{\textbf{6,837} $\pm$ 19}  & 6,561 \\ 
\multicolumn{1}{l|}{Humanoid-v3}    & 429 $\pm$ 20  & 582 $\pm$ 52  & N/A          & 5,499 $\pm$ 94    & 5,354 $\pm$ 337  & 5,758 $\pm$ 173 & \multicolumn{1}{c|}{\textbf{6,302} $\pm$ 136} & 5,855 \\ 
\multicolumn{1}{l|}{Walker2d-v3}    & 299 $\pm$ 75  & 554 $\pm$ 217  & 4,746 $\pm$ 316  & 3,391 $\pm$ 1,873  & 2,402 $\pm$ 949 & 6312 $\pm$ 47  & \multicolumn{1}{c|}{\textbf{6334} $\pm$ 10}  & 6,123 \\ 
\boldtoprule
\end{tabular}}
\vspace{-0.5em}
\end{table*}

%% file: Figures/training_curve.tex
\begin{figure}[t]
\includegraphics[width=\textwidth]{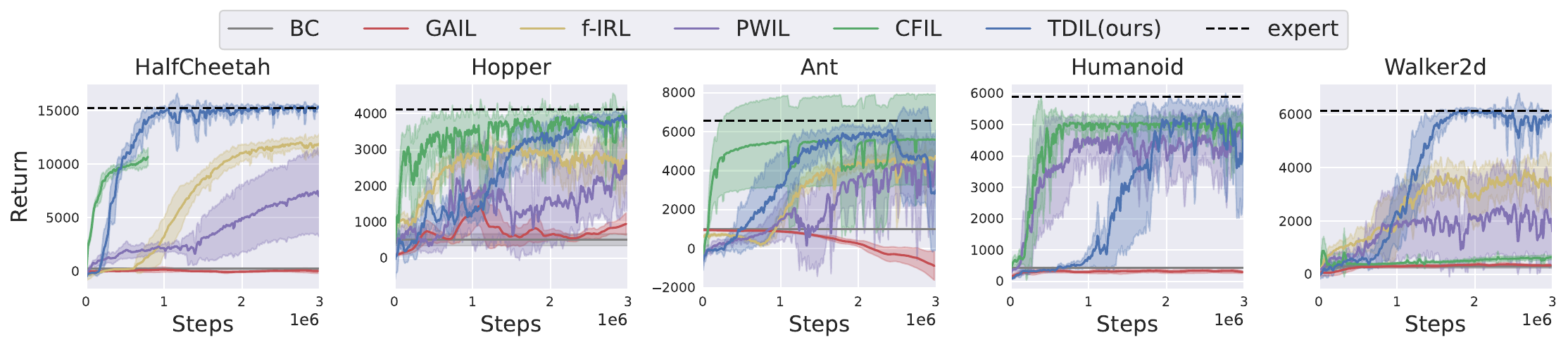}
\includegraphics[width=\textwidth]{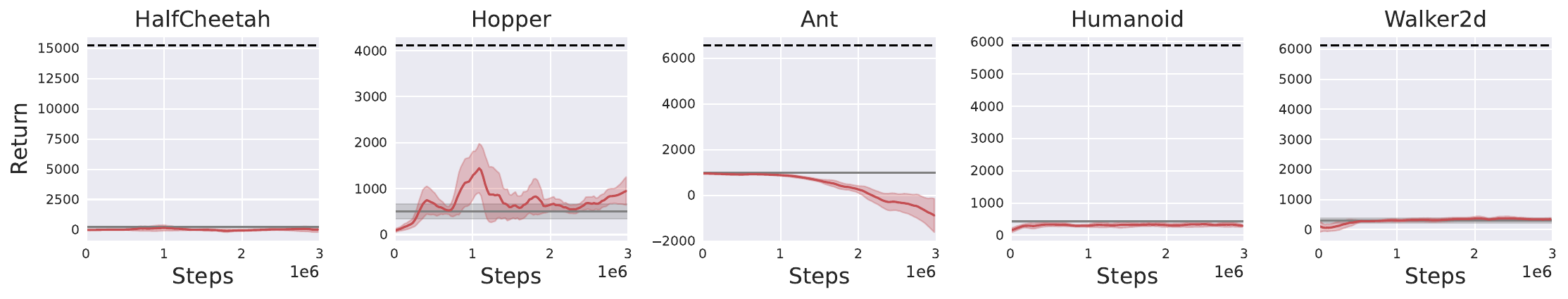}
\includegraphics[width=\textwidth]{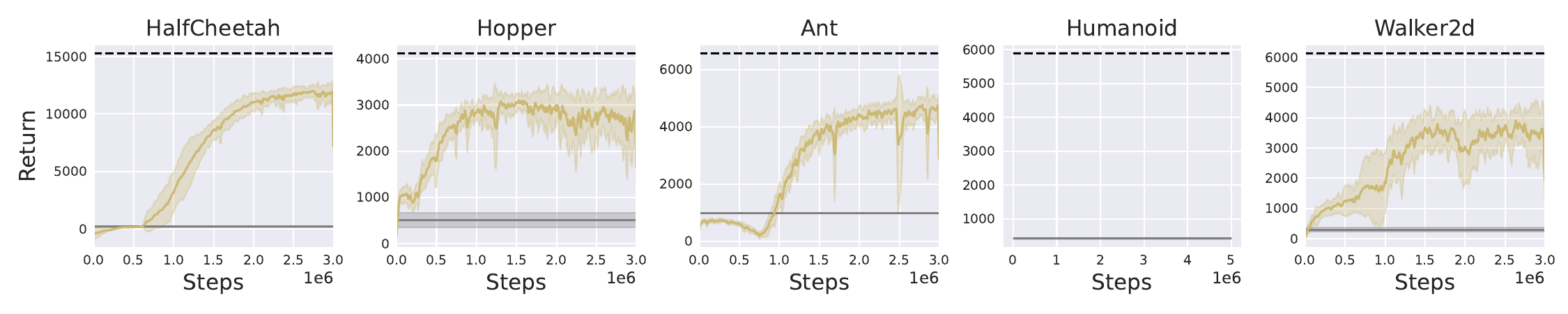}
\includegraphics[width=\textwidth]{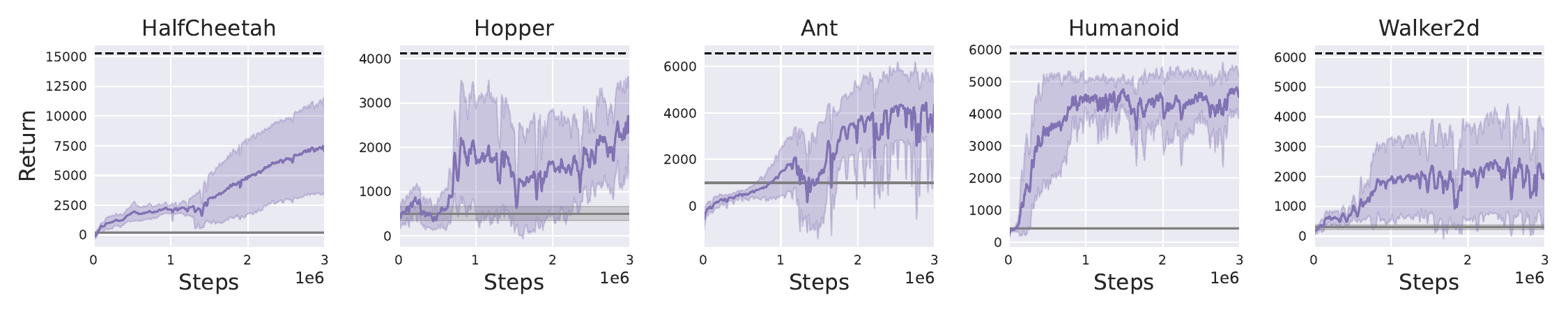}
\includegraphics[width=\textwidth]{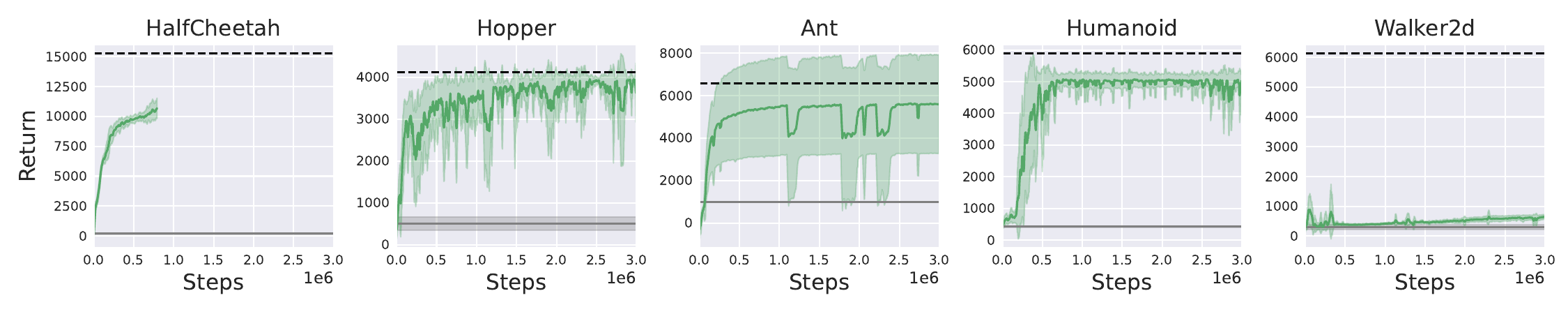}
\includegraphics[width=\textwidth]{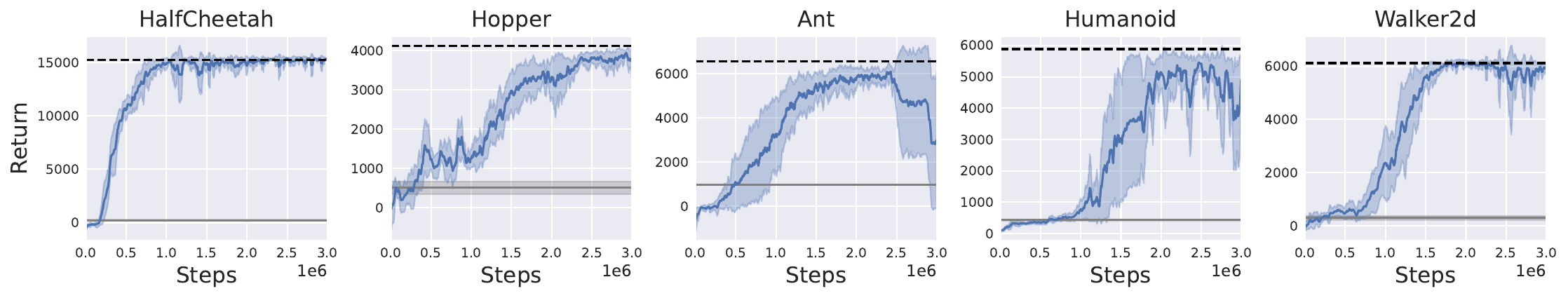}
\centering
\caption{The training curves of BC, GAIL, f-IRL, PWIL, CFIL, and TDIL. These curves represent the means and the standard deviations of five independent runs conducted with different random seeds}
\label{fig:training_curve}
\end{figure}

%% file: Table/multidemo.tex
\begin{table}[t]
\renewcommand{\arraystretch}{1.3}
\newcommand{\boldtoprule}{\midrule[1.5pt]}
\centering
\caption{TDIL when using multiple expert demonstrations}
\label{tab:multidemo}
\begin{adjustbox}{max width=\linewidth}
\small
\begin{tabular}{lccc}
\toprule
               & 1 Demo & 2 Demo & 3 Demo \\ \hline
\multicolumn{1}{l|}{HalfCheetah-v3} & 15,624 $\pm$ 119 (Expert: 15,251) & 15,711 $\pm$ 124 (Expert: 15,200) & 15,554 $\pm$ 293 (Expert: 15,197) \\
\multicolumn{1}{l|}{Hopper-v3}      & 4,115 $\pm$ 14 (Expert: 4,114) & 4,057 $\pm$ 65 (Expert: 4,194) & 4,068 $\pm$ 57 (Expert: 4,188) \\
\multicolumn{1}{l|}{Ant-v3}         & 6,837 $\pm$ 19 (Expert: 6,561) & 6,486 $\pm$ 308 (Expert: 6,417) & 6,341 $\pm$ 249 (Expert: 6,445) \\
\multicolumn{1}{l|}{Humanoid-v3}    & 6,302 $\pm$ 136 (Expert: 5,855) & 5,887 $\pm$ 308 (Expert: 5,926) & 6,035 $\pm$ 134 (Expert: 5,920) \\
\multicolumn{1}{l|}{Walker2d-v3}    & 6,334 $\pm$ 10 (Expert: 6,123) & 6,308 $\pm$ 52 (Expert: 6,095) & 6,252 $\pm$ 104 (Expert: 6,074) \\ \boldtoprule
\end{tabular}
\end{adjustbox}
\end{table}

%% file: Table/baselinebc.tex
\begin{table}
\renewcommand{\arraystretch}{1.3}
\newcommand{\boldtoprule}{\midrule[1.5pt]}
\centering
\caption{Performance of baselines with BC loss.}
\label{tab:baselinebc}
\begin{adjustbox}{max width=\linewidth}
\small
\begin{tabular}{ccccc}
\toprule
               & CFIL w/ BC  & PWIL w/ BC   & f-IRL w/ BC   & TDIL w/ BC \\ \hline
\multicolumn{1}{l|}{HalfCheetah-v3} & 14,853 & 4,679  & 13,638 & 15,666 \\ 
\multicolumn{1}{l|}{Ant-v3}         & 4,683  & 5,925  & 5,337  & 6,434  \\ 
\multicolumn{1}{l|}{Humanoid-v3}    & 5,343  & 5,294  & N/A   & 5,758  \\ 
\multicolumn{1}{l|}{Walker2d-v3}    & 6,286  & 5,489  & 4,403  & 6,312  \\ \boldtoprule
\end{tabular}
\end{adjustbox}
\end{table}

%% file: Figures/adroit.tex
\begin{figure}
  \centering
  \includegraphics[width=0.5\linewidth]{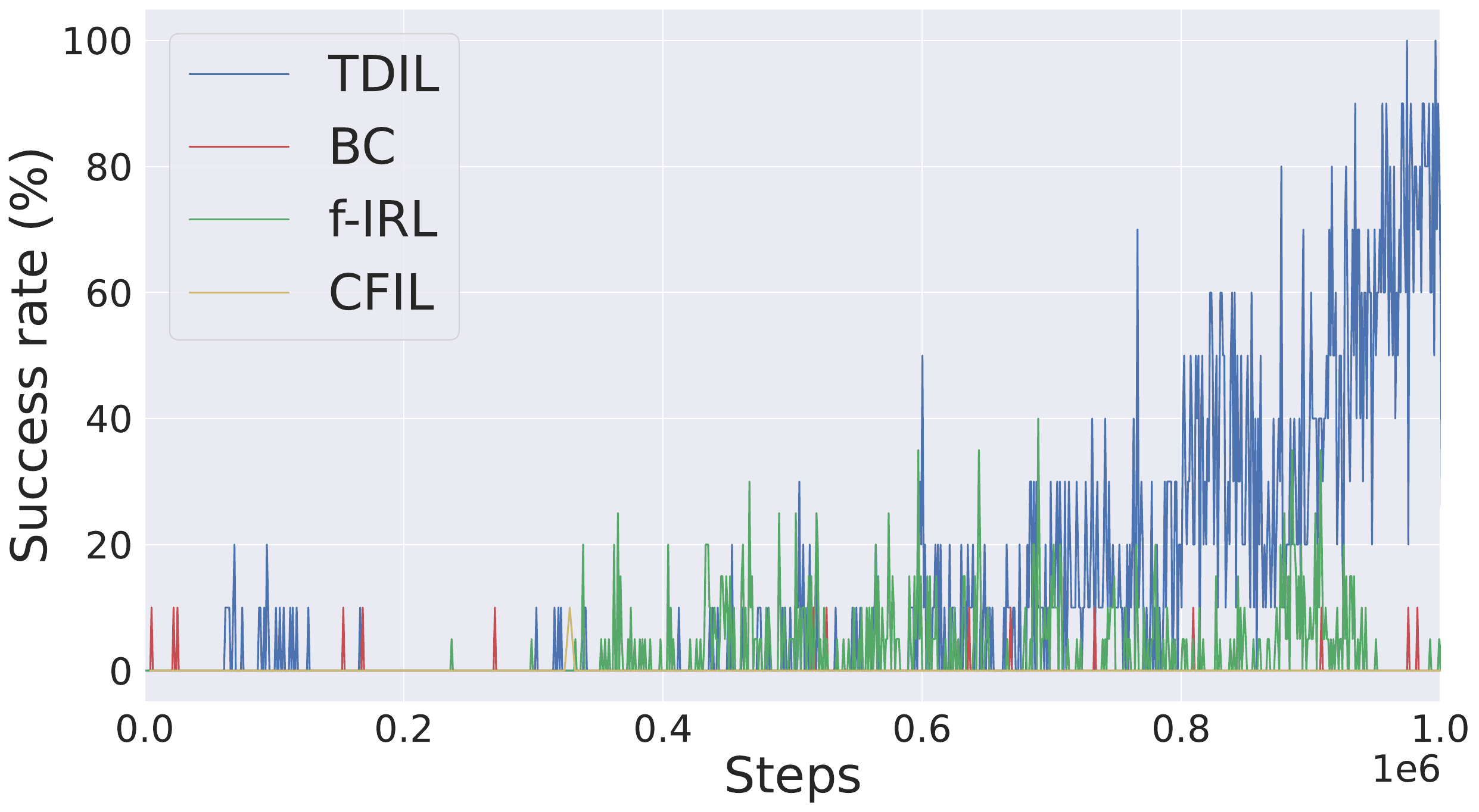}
  \caption{Comparing the success rates of TDIL, BC, f-IRL and CFIL in the AdroitHandDoor-v1 environment.}
  \label{fig:adroit}
\end{figure}

%% file: Figures/blind.tex
\begin{figure}[t]
\begin{subfigure}{0.9\linewidth}
\includegraphics[width=0.9\linewidth]{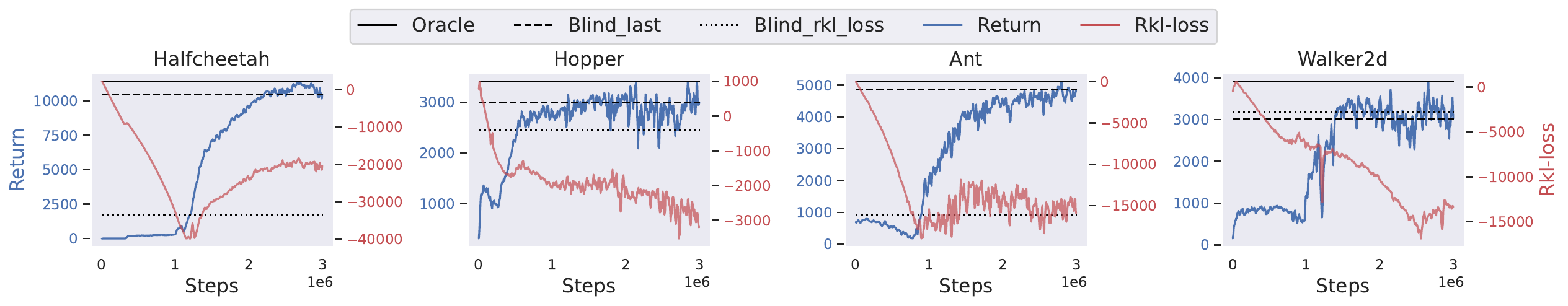}
\vspace{-0.5em}
\caption{f-IRL}
\vspace{1em}
\end{subfigure}
\begin{subfigure}{0.9\linewidth}
\includegraphics[width=0.9\linewidth]{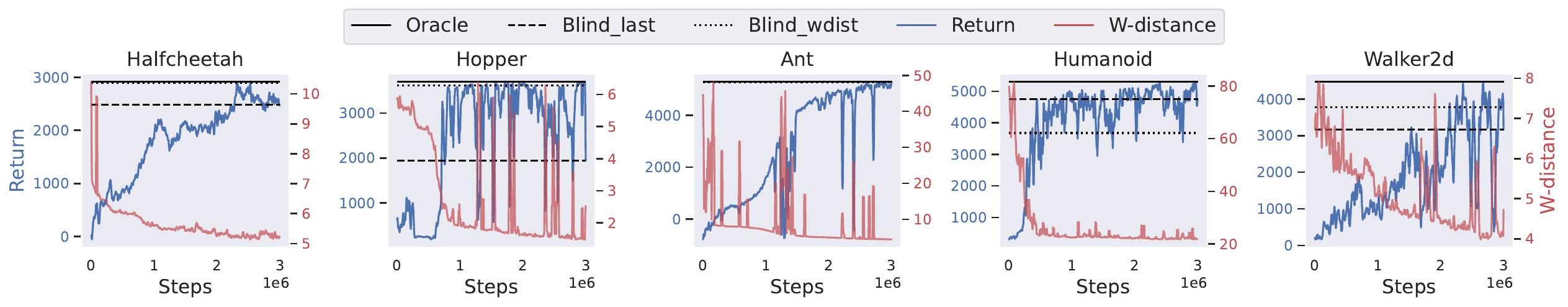}
\vspace{-0.5em}
\caption{PWIL}
\vspace{1em}
\end{subfigure}
\begin{subfigure}{0.9\linewidth}
\includegraphics[width=0.9\linewidth]{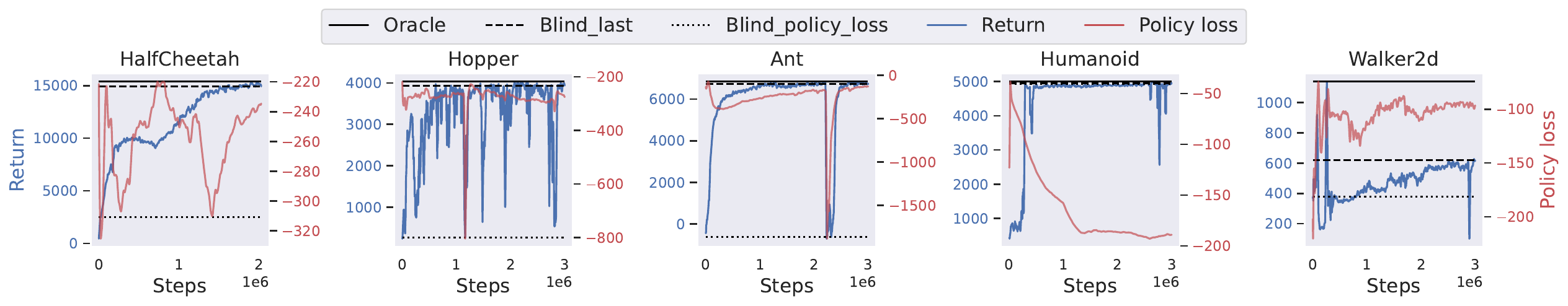}
\vspace{-0.5em}
\caption{CFIL}
\vspace{1em}
\end{subfigure}
\begin{subfigure}{0.9\linewidth}
\includegraphics[width=0.9\linewidth]{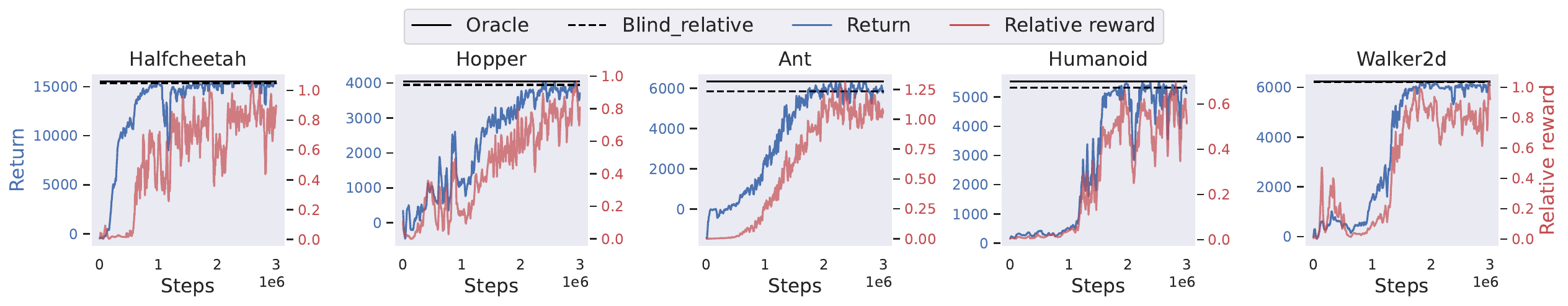}
\vspace{-0.5em}
\caption{TDIL~(ours)}
\vspace{1em}
\end{subfigure}
\centering
\caption{
The training curves and the blind selection results of f-IRL, PWIL, CFIL, and TDIL~(ours). The oracle line represents the highest evaluation return achieved during training. The Blind\_last line depicts the evaluation return achieved by the agent at the end of the training phase. The Blind\_\{rkl\_loss, wdist, policy\_loss, relative\} lines correspond to the evaluation returns determined based on the reverse KL loss, W-distance, policy-loss, and our proposed relative reward, respectively.
}
\label{fig:blind}
\end{figure}

%% file: Table/main_blind.tex
\begin{table}[t]
\renewcommand{\arraystretch}{1.3}
\newcommand{\boldtoprule}{\midrule[1.5pt]}
\vspace{0.25em}
\centering
\caption{Performance decrease ratios of different methods in the blind model selection scenario.
}
\label{tab:main_blind}
\resizebox{\linewidth}{!}{
\begin{tabular}{ccccccc}
\boldtoprule
                                    & BC~\citep{pomerleau1991efficient}         & f-IRL~\citep{firl2020corl}         & PWIL~\citep{dadashi2021primal}          & CFIL~\citep{freund2023coupled}          & Ours                               \\ \hline
\multicolumn{1}{l|}{HalfCheetah-v3} & -0.75 $\pm$ 0.34   & -0.27 $\pm$ 0.27 & -0.17 $\pm$ 0.15 & -0.07 $\pm$ 0.00 & \textbf{-0.02} $\pm$ 0.01 \\ 
\multicolumn{1}{l|}{Hopper-v3}      & -0.27 $\pm$ 0.23 & -0.32 $\pm$ 0.26   & -0.31 $\pm$ 0.29   & \textbf{-0.04} $\pm$ 0.02    & \textbf{-0.04} $\pm$ 0.05  \\ 
\multicolumn{1}{l|}{Ant-v3}         & -0.73 $\pm$ 0.03  & -0.18 $\pm$ 0.16   & -0.13 $\pm$ 0.18   & \textbf{-0.03} $\pm$ 0.01  & \textbf{-0.03} $\pm$ 0.01   \\ 
\multicolumn{1}{l|}{Humanoid-v3}    & -0.18 $\pm$ 0.16   & N/A           & -0.18 $\pm$ 0.16   & \textbf{-0.03} $\pm$ 0.03   & -0.04 $\pm$ 0.03    \\ 
\multicolumn{1}{l|}{Walker2d-v3}    & -0.45 $\pm$ 0.07  & -0.92 $\pm$ 0.08    & -0.45 $\pm$ 0.26  & -0.50 $\pm$ 0.31    & \textbf{-0.03} $\pm$ 0.04   \\ \boldtoprule
\end{tabular}}
\end{table}

%% file: Figures/accuracy.tex
\begin{figure}[t]
\begin{subfigure}{\textwidth}
\includegraphics[width=\textwidth]{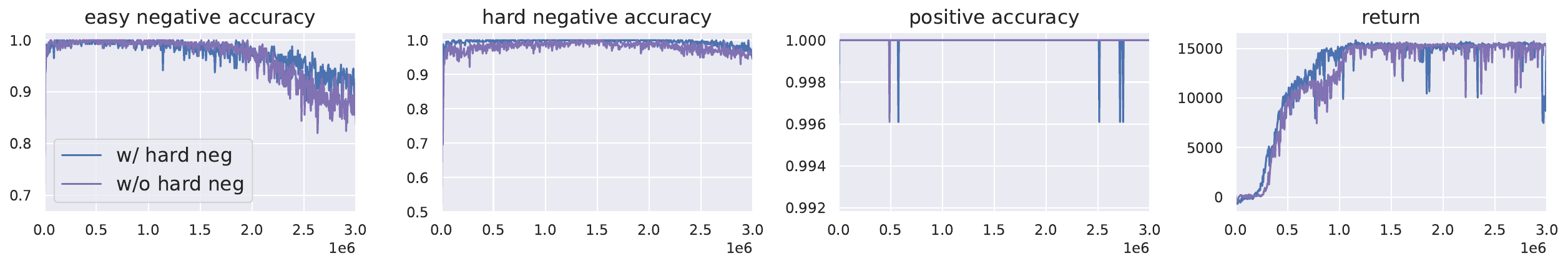}
\vspace{-1.5em}
\caption{HalfCheetah}
\vspace{1em}

\end{subfigure}
\begin{subfigure}{\textwidth}
\includegraphics[width=\textwidth]{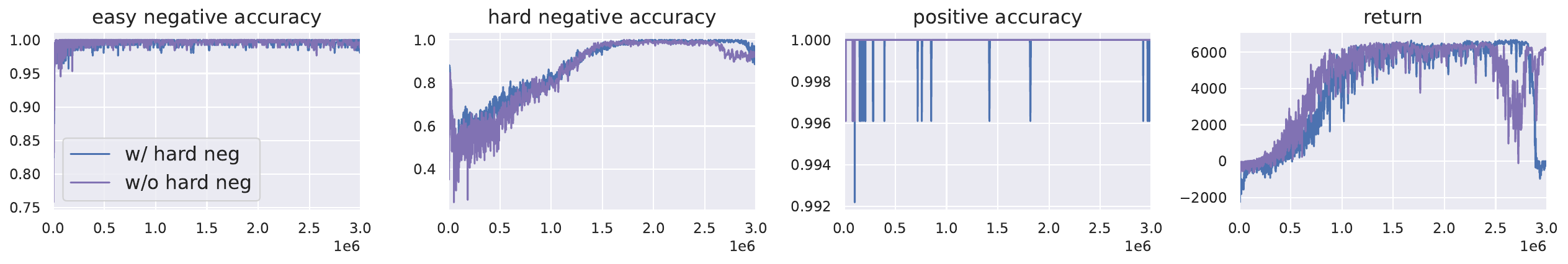}
\vspace{-1.5em}
\caption{Ant}
\vspace{1em}
\end{subfigure}

\begin{subfigure}{\textwidth}
\includegraphics[width=\textwidth]{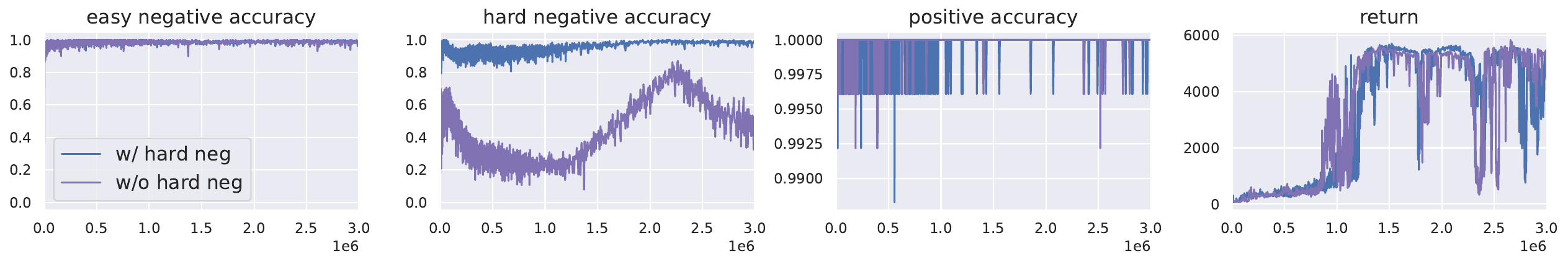}
\vspace{-1.5em}
\caption{Humanoid}
\vspace{1em}
\end{subfigure}

\centering
\caption{
An analysis of the accuracy of the transition discriminator in different environments
}
\label{fig:acc_td}
\end{figure}

%% file: Figures/alpha.tex
\begin{figure}
  \centering
  \includegraphics[width=\linewidth]{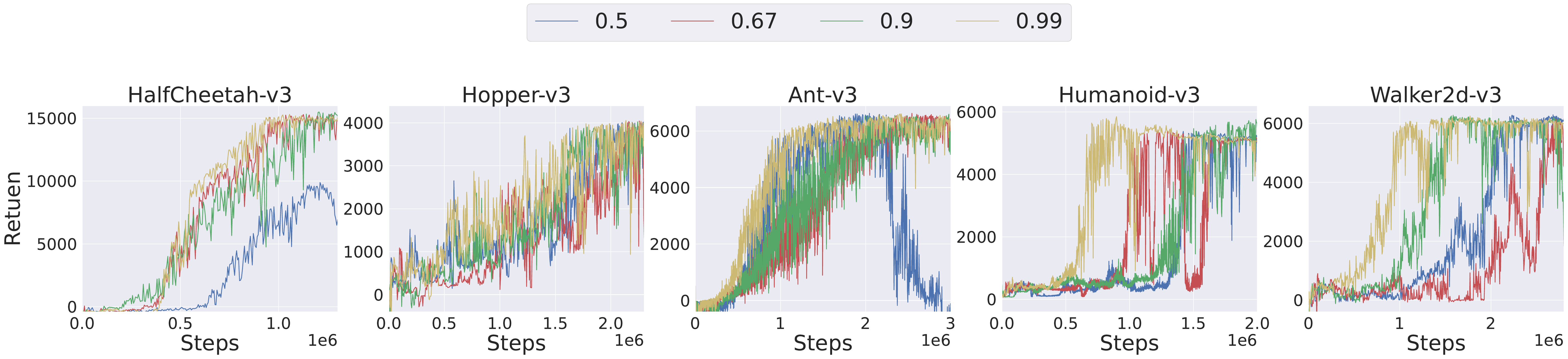}
  \caption{Performance of different $\alpha$ selection}
\label{fig:alpha}
\end{figure}

%% file: Table/tdacc.tex
\begin{table}
\renewcommand{\arraystretch}{1.3}
\newcommand{\boldtoprule}{\midrule[1.5pt]}
\vspace{0.25em}
\centering
\caption{Accuracy of the transition discriminator when trained with different $\alpha$. ``p" stands for positive data's accuracy; ``en" stands for easy negative data's accuracy; ``hn" stands for hard negative data's accuracy;}
\label{tab:tdacc}
\resizebox{\linewidth}{!}{
\begin{tabular}{ccccccccccccc}
\boldtoprule
               & 0.5(p) & 0.5(en) & 0.5(hn) & 0.67(p) & 0.67(en) & 0.67(hn)  & 0.9(p) & 0.9(en) & 0.9(hn) & 0.99(p) & 0.99(en) & 0.99(hn) \\ \hline
\multicolumn{1}{l|}{HalfCheetah-v3} & 1.0 & 0.997  & 0.998 & 1.0 & 0.996 & 1.0 & 1.0 & 0.997 & 0.996 & 1.0 & 0.992 & 0.996  \\ 
\multicolumn{1}{l|}{Hopper-v3} & 0.996 & 0.996  & 0.992 & 0.996 & 0.996 & 0.996  & 0.996 & 0.996 & 0.992 & 1.0  & 0.996 & 0.996 \\ 
\multicolumn{1}{l|}{Ant-v3}         & 0.996  & 1.0  & 1.0  & 1.0  & 1.0 & 1.0  & 1.0 & 0.996 & 1.0 & 1.0  & 0.992 & 0.992 \\ 
\multicolumn{1}{l|}{Humanoid-v3}    & 1.0  & 1.0  & 1.0   & 1.0  & 0.996 & 0.996  & 0.996 & 0.996 & 0.996 & 1.0  & 0.99 & 0.988 \\ 
\multicolumn{1}{l|}{Walker2d-v3}    & 0.996  & 0.996  & 0.996  & 0.992  & 0.996 & 0.996  & 0.996 & 0.992 & 0.99 & 1.0  & 0.992 & 0.988 \\ \boldtoprule
\end{tabular}}
\end{table}

%% file: Table/beta.tex
\begin{table}
\renewcommand{\arraystretch}{1.3}
\newcommand{\boldtoprule}{\midrule[1.5pt]}

\vspace{0.25em}
\centering
\caption{Performance of TDIL under different $\beta$ value selection.}
\label{tab:beta}
\resizebox{\linewidth}{!}{
\begin{tabular}{cccccccccccc}
\boldtoprule
               & 0+BC  & 0   & 0.1   & 0.2   & 0.5   & 0.8   & 0.9   & 0.95  & 0.99  & 1.0 & Expert \\ \hline
\multicolumn{1}{l|}{HalfCheetah-v3} & \textbf{15,666} & 12,630 & 15,100 & 15,541 & 15,479 & 15,612 & 15,624 & 15,462 & 15,529 & \multicolumn{1}{c|}{9,791} & 15,251\\ 
\multicolumn{1}{l|}{Hopper-v3}      & 4,115  & 3,890  & 4,124  & 4,126  & \textbf{4,162}  & 4,128  & 4,115  & 1,887  & 3,232  & \multicolumn{1}{c|}{1,950} & 4,114\\ 
\multicolumn{1}{l|}{Ant-v3}         & 6,434  & 3,995  & 6,358  & 6,513  & 6,467  & 6,611  & \textbf{6,837}  & 6,560  & 6,506  & \multicolumn{1}{c|}{4,216} & 6,561\\ 
\multicolumn{1}{l|}{Humanoid-v3}    & 5,758  & 5,575  & 6,288  & \textbf{6,352}  & 6,312  & 6,325  & 6,302  & 5,703  & 5,235  & \multicolumn{1}{c|}{1,826} & 5,855\\ 
\multicolumn{1}{l|}{Walker2d-v3}    & 6,312  & 6,281  & 6,251  & 6,204  & 6,266  & \textbf{6,346}  &  6,334  & 6,296  & 6,098  & \multicolumn{1}{c|}{1,769} & 6,123\\ \boldtoprule
\end{tabular}}
\end{table}

%% file: Sections/a5_multi_step_expert_proximity.tex
\subsection{Multi-Step Expert Proximity}
\label{apx:multi_step_expert_proximity}
In the main manuscript, the expert reachability indicator $\K_t$ is only defined to consider the transition to expert states within a single timestep. We could generalize the reachability indicator to multiple timesteps by defining $\K_t^{(k)}$, where it determines whether the state $s_t$ can reach an expert state by selecting a series of $k$ actions.
Formally, we define the following:
\begin{equation}
p(\K_t^{(k)}|s_t,a_t) \defeq
\begin{cases}
\int_{\mathcal{S}} p(s_{t+1}|s_t,a_t) p(\O_{t+1}|s_{t+1}) ds_{t+1}  & \text{if $k=1$,} \\
\int_{\mathcal{S}} p(s_{t+1}|s_t,a_t) p(\K_{t+1}^{(k-1)}|s_{t+1}) ds_{t+1}  & \text{if $k\in\{2,\dots,T\}$,} \\
\end{cases}
\end{equation}
The value of $p(\K_t^{(k)}|s_t)$ can be calculated as in the main manuscript. The surrogate reward functions corresponding to the indicators are defined as follows:
\begin{equation}
\begin{aligned}
\Rsur^{(k)}(s_t,a_t) &\defeq \mathbb{E}_{s_{t+1}\sim p(s_{t+1}|s_t,a_t)}\Big[p(\K_{t+1}^{(k)}|s_{t+1}) \Big].
\end{aligned}
\end{equation}
Each surrogate reward functions can be approximated by $\Dphistar^{(k)}(\sa,\sb)$ defined as:
    \begin{equation}
    \begin{aligned}
    \Dphistar^{(k)}(\sa,\sb) &\defeq \max_{a_i,\dots,a_{i+k-1}}\1\left[\prod_{j=i}^{i+k-1}P(\sb|\sa,a_j) > 0\right].
    \end{aligned}
    \label{eq:d}
    \end{equation}
The total reward function $\Rtot$ can then be re-defined based on the weighted sum of the surroagte reward functions $\Rsur^{(k)}$ across all $k$.